\documentclass[lettersize,journal]{IEEEtran}
\usepackage{amsmath,amsfonts}
\usepackage{algorithmic}
\usepackage{algorithm}
\usepackage{array}
\usepackage[caption=false,font=normalsize,labelfont=sf,textfont=sf]{subfig}
\usepackage{textcomp}
\usepackage{stfloats}
\usepackage{url}
\usepackage{verbatim}
\usepackage{graphicx}
\usepackage{cite}
\usepackage{hyperref}
\usepackage{pifont}
\usepackage{booktabs}
\usepackage{tikz}
\usepackage{makecell}
\usetikzlibrary{mindmap}
\usepackage[perpage,symbol*]{footmisc}
\DefineFNsymbols{circled}{{\ding{192}}{\ding{193}}{\ding{194}}
{\ding{195}}{\ding{196}}{\ding{197}}{\ding{198}}{\ding{199}}{\ding{200}}{\ding{201}}}
\setfnsymbol{circled}

\newcommand\myfootnotestyle[1]{\ifcase#1 \or \ding{182}\or \ding{183}\or
\ding{184}\or \ding{185}\or \ding{186}\or \ding{187}%
\or \ding{188}\or \ding{189}\or \ding{190}\or \ding{191}\else *\fi\relax}
\def\eg{\emph{e.g.}}
\def\ie{\emph{i.e.}}
\def\etc{\emph{etc}}

\def\etal{\emph{et.al.}}
\begin{document}

\title{Adversarial Examples in the Physical World:\\A Survey}
\author{Jiakai Wang, Xianglong Liu\textsuperscript{*}, Jin Hu, Donghua Wang, Siyang Wu, Tingsong Jiang, \\Yuanfang Guo,  Aishan Liu, and Jiantao Zhou
\thanks{J. Wang is with Zhongguancun Laboratory, Beijing, China.}
\thanks{D. Wang is with the College of Computer Science and Technology, Zhejiang University, Hangzhou, China.}
\thanks{X. Liu,  J. Hu, S. Wu, A. Liu, Y. Guo are with the State Key Lab of Software Development Environment, Beihang University, Beijing 100191, China.}
\thanks{T. Jiang and W. Yao are with the Defense Innovation Institute, Chinese Academy of Military Science, Beijing, China}
\thanks{J. Zhou is with the Department of Computer and Information Science, Faculty of Science and Technology, University of Macau, Macau, China}
}



\maketitle

\begin{abstract}
Deep neural networks (DNNs) have demonstrated high vulnerability to adversarial examples, raising broad security concerns about their applications. Besides the attacks in the digital world, the practical implications of adversarial examples in the physical world present significant challenges and safety concerns. However, current research on physical adversarial examples (PAEs) lacks a comprehensive understanding of their unique characteristics, leading to limited significance and understanding. In this paper, we address this gap by thoroughly examining the characteristics of PAEs within a practical workflow encompassing training, manufacturing, and re-sampling processes. By analyzing the links between physical adversarial attacks, we identify manufacturing and re-sampling as the primary sources of distinct attributes and particularities in PAEs. Leveraging this knowledge, we develop a comprehensive analysis and classification framework for PAEs based on their specific characteristics, covering over 100 studies on physical-world adversarial examples. Furthermore, we investigate defense strategies against PAEs and identify open challenges and opportunities for future research. We aim to provide a fresh, thorough, and systematic understanding of PAEs, thereby promoting the development of robust adversarial learning and its application in open-world scenarios\footnote{We construct a \href{https://github.com/jiakaiwangCN/awesome-physical-adversarial-examples}{sustainable open source project} to provide the community with a continuously updated list of physical world adversarial sample resources, including papers, code, \etc, within the proposed framework}. 
\end{abstract}


\section{Introduction}

The deep learning has achieved great success in various applications, such as computer vision (CV), natural language processing (NLP), and audio speech recognition (ASR). However, the vulnerability of deep learning models recently attracted great attention and gathered numerous researchers to investigate this critical issue. More precisely, the deep models are proven vulnerable to a kind of special-designed examples, which are termed adversarial examples \cite{DBLP:journals/corr/GoodfellowSS14}, that have small characterization differences with benign ones yet could easily fool deep models to output the wrong predictions. The adversarial examples were first proposed in the computer vision tasks in the digital world by Szegedy \etal~\cite{DBLP:journals/corr/SzegedyZSBEGF13} in 2014 while it was soon proven that be also effective in the physical world by Kurakin \etal~\cite{DBLP:conf/iclr/KurakinGB17a} in 2017. In the last five years, a series of works have strongly supported that the boom of physical adversarial examples poses great challenges to the security and safety of deep learning models in real systems. For example, Eykholt \etal~proposed the robust physical-world perturbation for the road sign recognition task, showing the great threats to real auto-driving correlated applications for the first time \cite{eykholt2018robust}. Liu \etal{} generated the trademark-like adversarial patches to attack the commodity identification models, which reveal the potential economic risks from vision-based automatic check-out systems \cite{DBLP:conf/eccv/LiuWLCZY20}. Wang \etal{} proposed an adversarial camouflage generation strategy in the 3D environment to mislead the vehicle classification and detection models, drawing more attention to the security of auto-driving system \cite{DBLP:conf/cvpr/WangLYLTL21}. In summary, the challenges of physical adversarial attacks have to be focused more on due to the trustworthiness requirement of deep learning based artificial intelligence.

On the mentioned basic, a great number of efforts have been made in constructing the in-depth understanding of the physical world adversarial examples by surveying in fashion physical attack studies~\cite{DBLP:journals/cybersec/SunTZ18,DBLP:journals/mlc/RenHY21,DBLP:journals/access/AkhtarMKS21,DBLP:journals/corr/abs-2211-01671,wang2022survey}. 
Although showing some benefits, these studies could not answer the critical question of \textit{what makes physical adversarial examples different from digital ones}.
Generally, physical adversarial examples show significant differences compared with digital adversarial examples in several aspects, such as characterization, generating strategy, and attacking ability. Yet substantially, the key particularity is that there is a line between their action environments, \ie, the digital-physical domain gap. To be specific, the physical world is a complex and open environment, where it has several dynamics such as lighting, natural noises, and diverse transformations. Thus, if adversaries expect to transfer digital-world adversarial examples into physical ones, they ought to overcome such varieties by adopting tailored efforts.
\IEEEpubidadjcol

Building upon the above insights, we are motivated to picture a more distinct hierarchy of physical world adversarial example generation methods, therefore promoting the understanding, development, and exploitation of physical examples. 
Specifically, we first revisit the critical particularities of physical adversarial examples under the perspective of workflow and give in-depth analysis in turn to induce the typical processes that might pose a great influence on adversarial examples generation. Briefly, we point out that performing physical adversarial attacks depends on the three processes, namely adversarial example optimization process, adversarial example manufacturing process, and adversarial examples re-sampling process, where the last two processes are specific to the physical adversarial attacks.
Furthermore, we give a 3-layer classification system to categorize the PAEs correlated studies based on the understanding of their special characteristics and critical attributes. According to the hundreds of physical world attack studies, we draw the route map of the PAE works and highlight the milestone studies.
Besides, we analyze the challenges of the PAEs from a concluded route map of the recent physical world adversarial attack studies for pushing its development. 
Also, we summarize the opportunities of the future PAEs to help protect applications from the physical world adversarial examples from the perspective of social good. We believe the conclusions could not only benefit the physical world adversarial examples but also show values in the whole of trustworthy artificial intelligence. 

To sum up, we believe that the proposed hierarchy in this paper could afford more clarity to the development of the PAEs. Based on the hierarchy, we provide a comprehensive understanding of the existing PAEs and a blue picture of the long-term studies about this hot area. Our combed defense strategies and challenges could give a greater awareness of practical deep learning in the future. 

The rest of this paper is organized as follows: We elaborate on the in-depth understanding of the PAEs in Section II, including the key particularities and critical attributes of PAEs. We classify the PAEs on the basis of the aforementioned hierarchy in Section III, including the manufacturing process-oriented PAEs, the re-sampling process-oriented PAEs, and others. Moreover, the defending strategies, the challenges, and the opportunities for confronting PAEs are summarized in Section IV. Finally, we conclude this work in Section V.

\section{Go Deep into PAEs}

\begin{figure}
\centering
\includegraphics[width=0.99\linewidth]{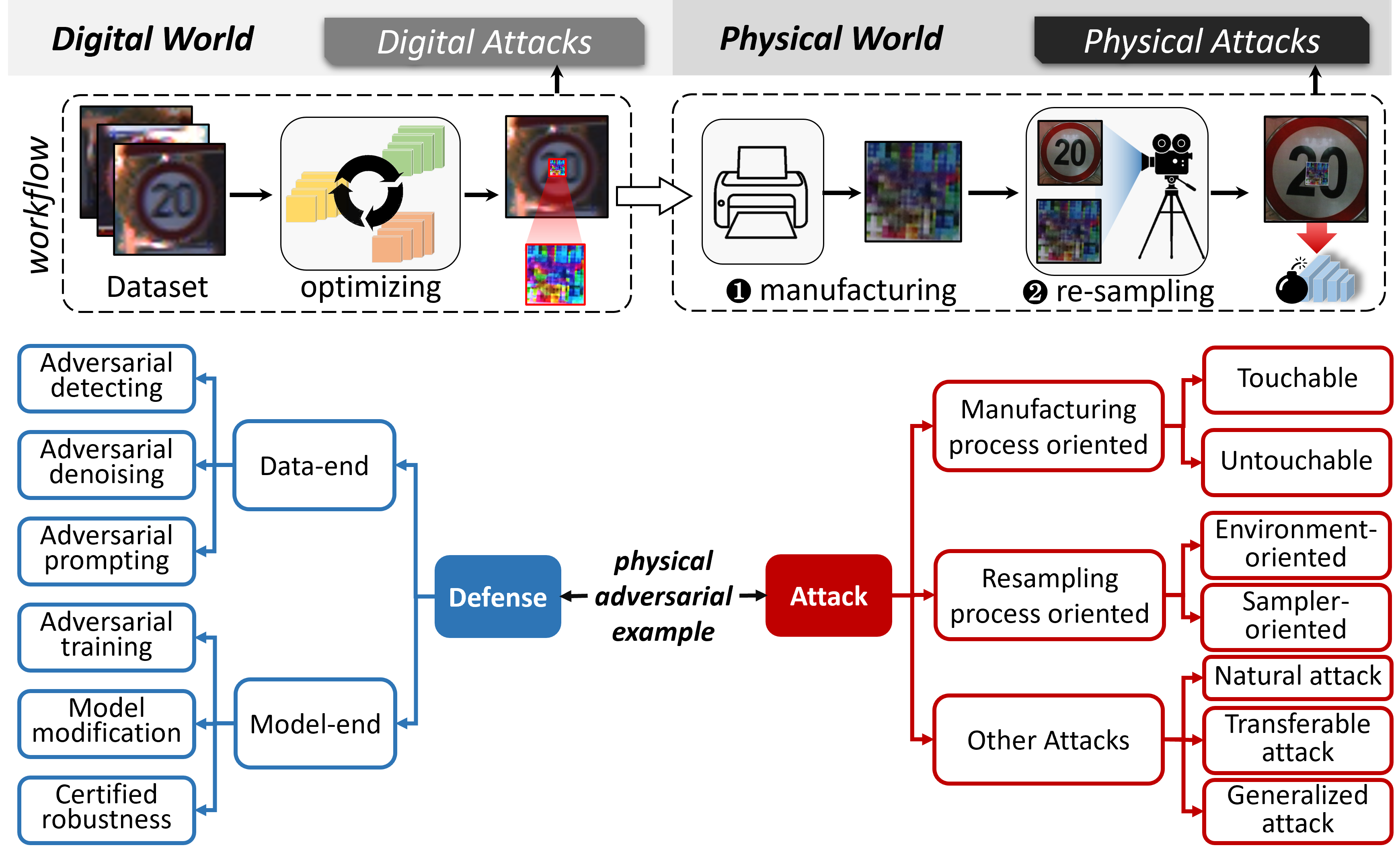}
\caption{The differences between digital and physical world examples via the perspective of generation workflow.}
\label{fig:phy-dig-diff}
\end{figure}
\subsection{Overview}
To build an appropriate hierarchy for physical world adversarial examples, it is necessary to go deep into the essential differences between digital and physical world examples. As we all know, the targeted action environments, \ie, the digital world and physical world, divide the adversarial examples into digital and physical kinds. Considering the process of becoming effective, it is much more complex to generate PAEs due to the necessary additional process compared with digital ones. Roughly speaking, the physical world adversarial examples could perform strong adversarial attacks in the digital world, while the digital world adversarial examples do not consistently maintain acceptable attacking ability in the physical world. 

As shown in Figure \ref{fig:phy-dig-diff}, taking a computer vision scenario as a typical instance, we give a workflow-based diagram to discriminate the essential distinctions between digital and physical world examples. It can be witnessed that digital and physical world examples have similarities and differences in generation workflow. 
Precisely, they share similar adversarial noise optimizing processes, which allows the adversarial noises (\eg, perturbations, patches, and so on) to acquire the adversarial attacking ability.
However, the key processes that make the PAEs distinctive are the adversarial pattern manufacturing process and the adversarial examples re-sampling process. 
During the manufacturing process, the adversaries manufacture the adversarial patterns via various approaches, such as producing material objects and playing with diverse projection devices or audio devices. For computer vision tasks, the output of this process could be adversarial patches \cite{doan2022tnt}, adversarial hats \cite{komkov2021advhat}, adversarial T-shirts \cite{xu2020adversarial}, adversarial lighting \cite{sugawara2020light}, and \etc. For audio recognition tasks, the output of this process could be adversarial audio music \cite{li2019adversarial}, adversarial audio noise \cite{yakura2019robust}, and \etc. 
During the re-sampling process, the manufactured adversarial physical patterns will be put into real scenarios, \ie, diverse environments, to act attacks in turn. These physical adversarial patterns could be sampled by different samplers as re-sampled digital adversarial patterns and make efforts to mislead or fool deep learning model-based intelligent equipment in the real world. Specifically, the samplers in this process could be cameras, tape recorders, video recorders, infrared imagers, smartphones, \etc. 

\subsection{The Key Particularities among PAEs}
Building upon the given overview of the physical adversarial examples, we go further to discuss the key particularities among them.
As we mentioned above, what makes the PAEs different from the digital ones are the particular generation processes, \ie, the manufacturing process and the re-sampling process. Hence, it is reasonable for us to explore the particularities through the two special steps. 

Regarding the manufacturing process, it mainly aims to manufacture the digitally-trained adversarial patterns into the physical environment of existing objects, which indicates a ``virtual-to-real'' process. The key particularities correlated to performing PAEs in manufacturing process are the \textbf{adversarial pattern manufactured techniques} and \textbf{adversarial pattern manufactured carriers}. For example, an image classification-oriented physical adversarial patch generation method should consider the two particularities. Manufacturing the patch-like physical adversarial patterns with a color high-resolution printer will achieve stronger attacking ability than a low-resolution printer, because the high-resolution printer could preserve much more detailed patterns that have attacking ability. As for the manufactured carriers, taking the art paper as a print carrier might achieve better performance than taking the whiteboard paper as a print carrier because the art paper has a smoother surface and better inking performance, which benefits the preservation of adversarial patterns. Similar particularities can also be found in audio recognition tasks, like the different parsers and renders in the players.

Regarding the re-sampling process, it re-samples the manufactured PAEs from the real environment as the output, which is in fact a ``real-to-virtual'' process. The key particularities correlated to conducting adversarial attacks during this process are also shown in two aspects, namely, the \textbf{sampling environment} and the \textbf{sampler quality}. More precisely, in different sampling environments, there always exist diverse environmental conditions that cause noises, such as illumination, weather, blur, visual angle, distance, external noise, and \etc. These kinds of noises commonly exist in the physical world and might cause the attacking ability to lose in some cases. The sampler quality also plays an important role in PAE generation because of the sampling precision deviations. Taking images as an instance, a Cannon EOS 5D sampler that has 30.4 million effective pixels could sample the PAEs with lower pattern losses than a Cannon EOS 80D sampler that has 24.2 million effective pixels, the same in the audio tasks.

In summary, the concluded key particularities among PAEs can be described as a ``virtual-real-virtual'' process, which is mainly devoted to overcoming the significant digital-physical domain gap (\ie, the loss of the adversarial patterns originated from digital-trained adversarial examples) between the digital and physical world. For acting in the real world, the adversaries have to manufacture the adversarial patterns into real objects. To attack the devices within deployed deep models, the physical adversarial patterns have to suffer the re-sampling operation. Therefore, the PAEs should always consider the possible impacts from the mentioned key particularities above, namely, the manufactured techniques, the manufactured carriers, the sampling environment, and the sampler quality. To be specific, these key particularities could result in significant drops in attacking ability by giving rise to the distortions of the adversarial patterns between PAEs and the trained adversarial examples. In practice, these concluded key particularities of PAEs are always the imperative factors that should be considered, which constructs the cornerstone of performing effective attacks against the intelligent devices in the physical world.

\subsection{The Critical Attributes of PAEs}
\begin{figure}
    \centering
    \includegraphics[width=0.9\linewidth]{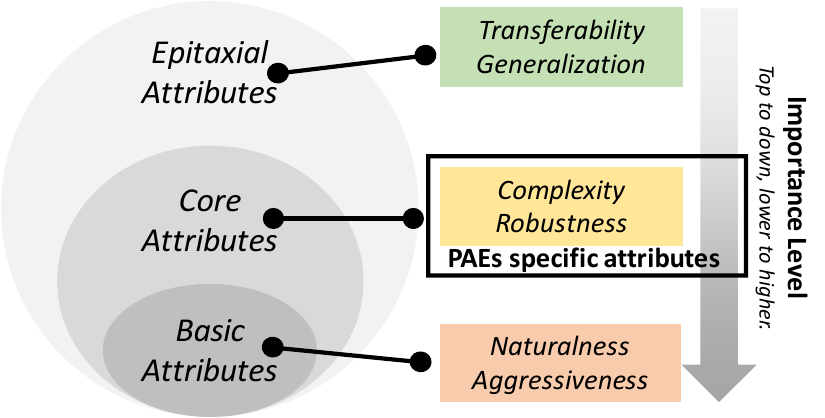}
    \caption{The Venn diagram of the relationship of the different kinds of critical attributes of the physical adversarial examples.}
    \label{fig:venn}
\end{figure}

Backend by the concluded particularities with full consideration of the physical-world natures. We further investigate the critical attributes of the PAEs.
In the open and real physical world, the adversarial examples might face \textit{complex guards}, \textit{dynamic environments}, and \textit{diverse targets}, making it necessary to take these physical challenges into account. Therefore, we summarize the critical attributes of PAEs by considering both the mentioned challenges and the particularities, finally dividing them into 3 categories in this paper. Specifically, they are basic attributes, core attributes, and epitaxial attributes. We give a brief Venn diagram to picture the relationship as shown in Figure \ref{fig:venn}. In the following sections, we will give a detailed description of the three kinds of attributes and interpret the reasons behind them.

When it comes to the critical attributes of physical adversarial examples, we have to take the first step from the typical adversarial examples, \ie, digital adversarial examples. As proposed in \cite{DBLP:journals/corr/GoodfellowSS14}, the adversarial example is defined as a kind of imperceptible example that could mislead deep models into wrong predictions. Hence, born from the digital world adversarial examples, the physical adversarial examples also possess the characteristics of digital ones, \ie, the naturalness and aggressiveness, which respectively represent the ability of ``imperceptible'' and ``attacking''. Since these two attributes originated from the primitive definition of adversarial examples, we thus call them \textbf{basic attributes}, which means that these are indispensable factors to be thought of when generating physical adversarial examples. Besides, these two attributes also satisfy the requirements of responding to the complex guards challenge. To be specific, the naturalness is useful in misleading human perception and the aggressiveness benefits attacking models, especially the enhanced robust models. However, it should be noted that the naturalness of the physical adversarial examples is quite different from that in the digital world. Specifically, the digital world naturalness constrains the adversarial noises with a very tiny $\epsilon$- norm, which makes the generated adversarial examples have nearly no difference from the appearance of the corresponding benign examples. While the physical adversarial examples allow the generated examples to show certain differences in appearance with relevantly looser constraints, \ie, perceptible but not conspicuous.

Furthermore, according to the particular process of physical adversarial examples, we summarize the attributes that are exclusive to physical adversarial examples as \textbf{core attributes}, which include complexity, and robustness. 
To be specific, the complexity evaluates the level of difficulty in manufacturing the physical adversarial examples. Since we have to produce the generated adversarial examples with different carriers and techniques in the real world, it is not desirable to design physical adversarial examples that are not easy to manufacture. In other words, once the manufacturing process is complex and unworkable, we think that it can not be treated as an awesome physical adversarial example generation method.
Besides, the complexity is also important for responding to the dynamic environments. To be specific, if a PAE is complex to generate, it might be difficult to successfully perform attacks in dynamic environmental conditions due to unsatisfactory reproducibility.
The robustness is defined as the degree of retaining the attacking ability when physical adversarial examples face different environmental noises. The reason why robustness is important to physical adversarial examples is evidenced by the fact that the re-sampling process always introduces various noises that do not exist in the digital world. If a physical adversarial example shows strict requirements of environment settings, it means its application range and practical value are limited. This attribute is also correlated to the dynamic environments in the real world.

Moreover, considering the real-world facts of employing physical adversarial examples, \ie, the diverse targets, we think the \textbf{epitaxial attributes} should also be taken into account. Different from digital adversarial examples, physical adversarial examples perform attacks in an open-set and complex real-world environment. Nevertheless, in such an environment, there exist multiple models and instances with diverse architectures and patterns, which challenges the attacking ability of the generated physical adversarial examples. Thus, the epitaxial attributes of physical adversarial examples mainly include transferability and generalization, which respectively concern the capability of attacking diverse models and diverse categories. It is easy to understand why transferability and generalization are necessary to be considered. However, these two important attributes play critical roles in not only physical world adversarial example generation, but also digital world adversarial example generation, and that is why we regard them as epitaxial attributes. We can observe that several existing digital adversarial attack methods achieve results such as \cite{Zhou_2018_ECCV,NEURIPS2021_7486cef2,moosavi2017universal,hendrik2017universal}. Despite being of significance in digital adversarial examples, we have to realize that the epitaxial attributes are much more meaningful to the physical adversarial examples with respect to the open and complex real world.


\subsection{The Definition of PAEs}

Based on the aforementioned analysis, we now give the definition of the PAEs in this section. First, we re-visit the source of the PAE, \ie, the digital adversarial examples, which is also called the ``adversarial example'' in common sense. The adversarial example $x_{adv}^d$ is first proposed by Szegedy \etal~\cite{DBLP:journals/corr/SzegedyZSBEGF13} and is described as a kind of examples that are obtained by imperceptibly small perturbations to a correctly classified input image $x$ but is able to mislead the deep model $\mathcal{F}$, \ie, $y^x \neq \mathcal{F}(x_{adv}^{d})$, $x_{adv}^{d} = x + \delta $, where $y^x$ is the ground-truth label of the input instance $x$, $\delta$ indicates the adversarial perturbation, and it satisfies $\lVert \delta \rVert < \varepsilon$ ($\varepsilon$ is a small enough radius and bigger than 0). Under such a definition, the digital adversarial perturbation always appears as a kind of ``noise-like'' image in computer vision. 
While in the physical world scenarios, we emphasize that the adversarial examples have to overcome the challenges, such as the information missing during manufacturing and re-sampling processes. It is difficult to constrain the adversarial perturbations to a very limited budget, such as a small $\varepsilon$ radius, due to the attacking ability degeneration risks. Therefore, taking computer vision tasks as instances, most of the PAE generation methods replace the ``noise-like'' perturbation with ``patch-like'' textures, which do not fully satisfy the original definition of the digital world. To unify the definition of the adversarial examples in both the digital and physical world, we here give a modified definition of the adversarial examples in the physical world:
\begin{equation}
\begin{split}
     y^x \neq \mathcal{F}(x_{adv}^p)&, \quad\emph{s.t.}, \quad\lVert x_{adv}^p\rVert _{\aleph} < \varepsilon, \\
    x_{adv}^p =& x + \mathcal{R}(\mathcal{M}(\delta), c),
\end{split} 
\label{equ:pae}
\end{equation}
where  $x_{adv}^p$ is the input physical adversarial example to the deployed deep models, $\mathcal{R}(\cdot)$ is the re-sampling function that represents the re-sampling process, $\mathcal{M}(\cdot)$ is the manufacturing function that represents the manufacturing process, $c$ is a certain environment condition and comes from the real and infinite environment conditions that are denoted as $\mathbb{C}$, \ie, $c \in \mathbb{C}$, the $\lVert\cdot\rVert_{\aleph}$ represents the evaluation metric that measures the naturalness of the PAE that input to the deployed artificial intelligence system, $\aleph$ indicates the recognizable space of human beings to the PAEs. To be brief, the $\aleph$ constraint imposed on the $\delta$ correlates to the ``suspicious'' extent of PAEs, making them acceptable for human beings in real scenarios.

\section{Classify Physical Adversarial Examples}

In this section, we focus on the key particularities of physical adversarial examples as outlined, including the manufacturing process and the re-sampling process, to outline the physical adversarial examples studies in the recent 6-year.
According to the \textit{No Free Lunch Theorem}~\cite{NoFreeLunch}, optimization algorithms that are effective in one specific domain may not be suitable for other domains. Adversarial example generation itself requires finding out-of-distribution (OOD) data from the perspective of deep learning models. Therefore, it is necessary to independently design methods for the manufacturing and resampling problems in different scenarios. Many research efforts have been undertaken in this regard, and we further classify these works based on specific scenarios.
Furthermore, in the subsection \textit{Others Attacks}, we provide a review of the cross-work of transferable, generalized, and on-manifold or natural adversarial examples, which have been the focus of recent discussions, in the context of physical adversarial examples.
Note that we provide elaborated tables to provide more collected details, like targeted tasks, authors, codes, publishers, and so on, of each kind of PAE in the appendix.

\subsection{Overview}

\begin{figure*}[t]
    \centering
    \includegraphics[width=.99\linewidth]{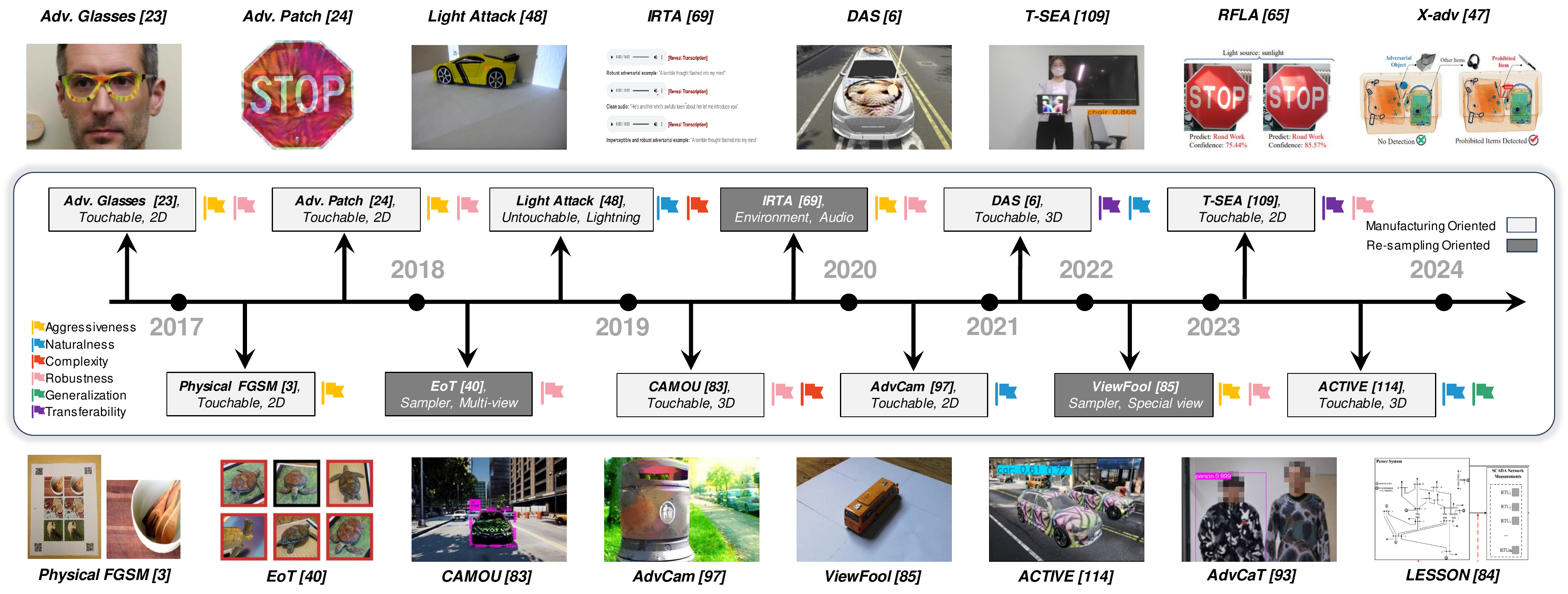}
    \caption{The route map of physical adversarial examples. This figure indicates the development tendency of the PAEs.}
    \label{fig:route_map}
\end{figure*}

To reflect the development veining, we build the route map of physical adversarial examples by discriminating the publishing year, the concerned process, the correlated attributes, and the targeting task of several typical and classical studies. The route map is shown in Figure \ref{fig:route_map}, where we can witness several meaningful conclusions. Briefly, we can draw such conclusions: (1) at the early stage, how to manufacture PAE is the key problem, and the re-sampling process-oriented PAE studies are now going into development; (2) the early PAE studies focus more on the basic attributes and core attributes, while the epitaxial attributes attract more attention in recent years; (3) overall, most of the PAE studies try to address mixture issues, \eg, DAS~\cite{DBLP:conf/cvpr/WangLYLTL21} try to generate natural, transferable physical adversarial camouflages with low complexity (training PAEs in 3D environment), which also indicates that the PAE research always faces composite problems due to the complex and open real-world scenarios.

\subsection{The Manufacturing Process Oriented PAEs}
As aforementioned, the first special process that makes physical attacks different from digital ones is the manufacturing process, which decides the conformation of the physical adversarial examples. As for categorizing the manufacturing process-oriented PAEs, there could be several principles, such as material-driven (categorizing the PAEs via the differences among materials),  and task-driven (categorizing the PAEs via the applied tasks). While considering the practicality of physical scenarios and scalability to future studies, we divide the manufacturing process-oriented PAEs into 2 categories that are significantly different in modeling, \ie, touchable attacks and untouchable attacks, where the former indicates that the generated adversarial examples could be touched by hands and the latter could not.

\begin{figure}[b]
    \centering
    \includegraphics[width=0.99\linewidth]{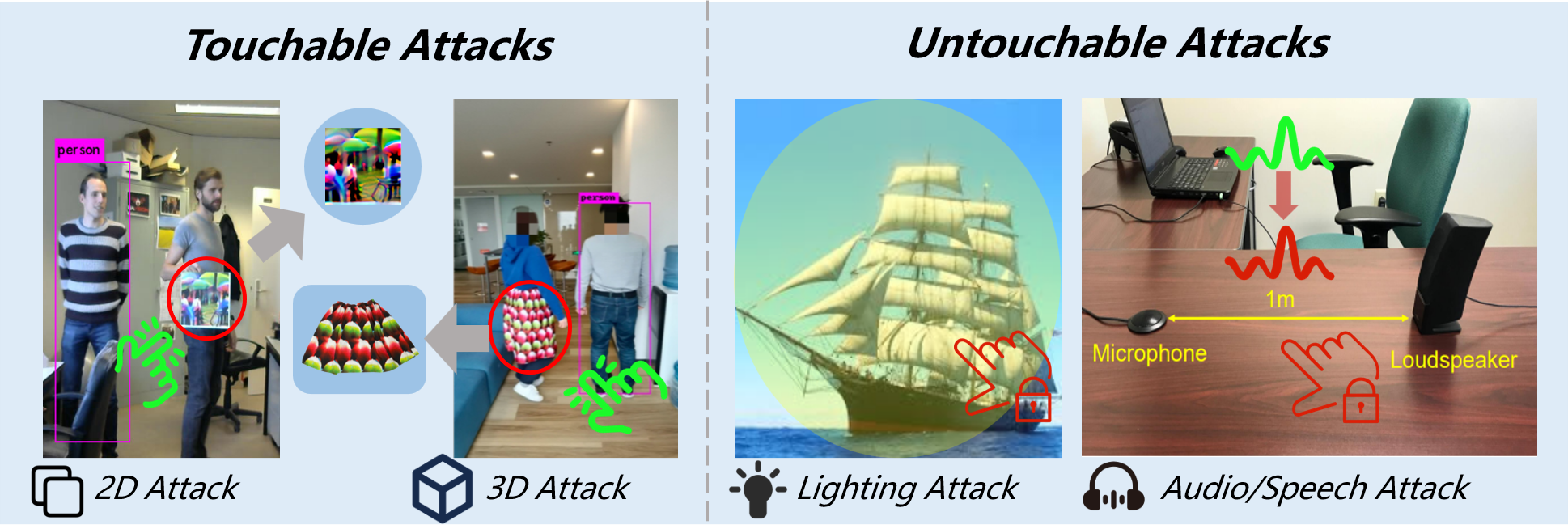}
    \caption{The overview of the manufacturing process oriented PAEs.}
    \label{fig:manufacture}
\end{figure}

\subsubsection{Touchable Attacks} The touchable attack includes 2D and 3D adversarial examples that respectively have 2D/3D forms, or consider the 2D/3D characteristics. In other words, this kind of PAEs could be generated by certain materials.

\ding{182} \emph{\textbf{2D attacks}.}
Sharif \etal~\cite{sharif2016accessorize} first investigated the vulnerability of face recognition systems toward physical adversarial attacks. They took the smoothness and practicability of physical attack into consideration and proposed to optimize the pattern of the eyeglass frame to deceive the face recognition system. To achieve smoothness, the author adopted the \textit{total variation} (TV)  loss. The TV loss quantifies the smoothness of the adversarial patch in the image, which can be formulated as:
\begin{equation}
\begin{split}
     \mathcal{L}_{tv} = \sum_{i,j} \sqrt{ {(p_{i+1,j} - p_{i,j})}^2  +  {(p_{i,j+1} - p_{i,j})}^2}.
\end{split} 
\label{equ:tv-loss}
\end{equation}
To realize practicability, they considered the color discrepancy between the digital color and the available color of the printer and devised \textit{non-printability score (NPS)} loss function to encourage the color of the optimized pattern to be printed out without loss. Let $ P \in {[0,1]}^3 $ denote the set of printable RGB triplets, then the NPS of a pixel $\hat{p}$ can be defined as:
\begin{equation}
\begin{split}
     \mathcal{L}_{NPS}=\prod_{p\in P} \lvert \hat{p}-p \rvert .
\end{split} 
\label{equ:nps}
\end{equation}

In the early stages of creating physical adversarial examples, the primary focus was on basic attributes of aggressiveness. As a result, the form of the PAEs obtained was relatively simple. Lu \etal~\cite{lu2017adversarial} demonstrated a minimization procedure to create adversarial examples that fool Faster RCNN in stop sign and face detection tasks. However, due to the restrictive environmental conditions, this adversarial attack did not perform well in the physical world. Kurakin \etal~\cite{ifgsm2018adversarial}  demonstrated the possibility of crafting adversarial examples in the physical world by simply manufacturing printout adversarial examples, re-sampling them with a cellphone camera, and then feeding them into an image classification model. The success of the attack sent a positive signal for the research of physical adversarial examples. Moreover, Eykholt \etal~\cite{eykholt2018robust} first generated adversarial perturbations in the physical world against road sign classifiers and proposed the Robust Physical Perturbations (RP2) algorithm. By optimizing and manufacturing white-black bock perturbation, the authors successfully attacked the traffic sign recognition model. Based on this work, Eykholt \etal~\cite{eykholt2018physical} extended the RP2 algorithm to object detection tasks and manufactured colorful adversarial stop sign posters. Further, Lee \etal~\cite{lee2019physical} first proposed an adversarial patch-attacking method that could successfully attack detectors without overlapping the target objects. Thys \etal~\cite{Thys2019fooling} first generated physical adversarial patches against pedestrian detectors by optimizing a combination of adversarial objectiveness loss, TV loss, and NPS loss.

Unlike previous works that modify adversarial examples in the perturbation process to meet additional objectives, Sharif \etal~\cite{sharif2019general} propose a general framework to generate diverse adversarial examples that meet the desired objectives. To achieve this, they construct adversarial generative nets(AGNs), which are flexible to accommodate various objectives, \emph{e.g.}, unconsciousness, robustness, and scalability. 



Researchers also shed light on the computational performance of generating PAEs. Wang \etal~\cite{wang2021daedalus}  proposed a subsampled non-printability score (SNPS) loss, which calculates a subsample pixel of the adversarial patch, accelerating the compute efficiency for the high-resolution patch (e.g., poster) without sacrificing attack performance. 
Wei \etal~\cite{wei2022adversarial} pointed out that the drawbacks of patch-based attack methods are susceptible to color distortion caused by the printer and proposed an attacking method based on the position and rotation angle. 


Recently, Wei \etal~\cite{wei2023hotcold} pointed out that adversarial patterns against the infrared object detector created by existing attack methods are conspicuous to humans. Thus, the author adopted the common cooling paste and warming paste as the media to manufacture physical adversarial examples. 
%

With the recent advancements and deployment of Deep Learning-Based Perception and Cognitive Systems, 2D adversarial examples have also demonstrated the ability to perform physical world attacks in new models.  Zhang \cite{zhang23CAPatch} extended the adversarial attack on object detectors to image caption systems, and evaluated the method in different model architecture, demonstrating  the image xaption system is also vulnerable to attacks using small adversarial patches. 

\begin{figure}[t]
	\centering
	\begin{minipage}{.3\linewidth}
		\centering
		\includegraphics[width =1\linewidth]{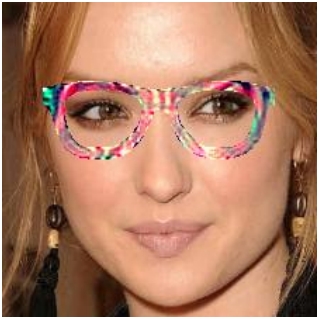}
		\centerline{\footnotesize  Sharif \etal~\cite{sharif2016accessorize}} 
	\end{minipage}
	\begin{minipage}{.3\linewidth}
		\centering
		\includegraphics[width =1\linewidth]{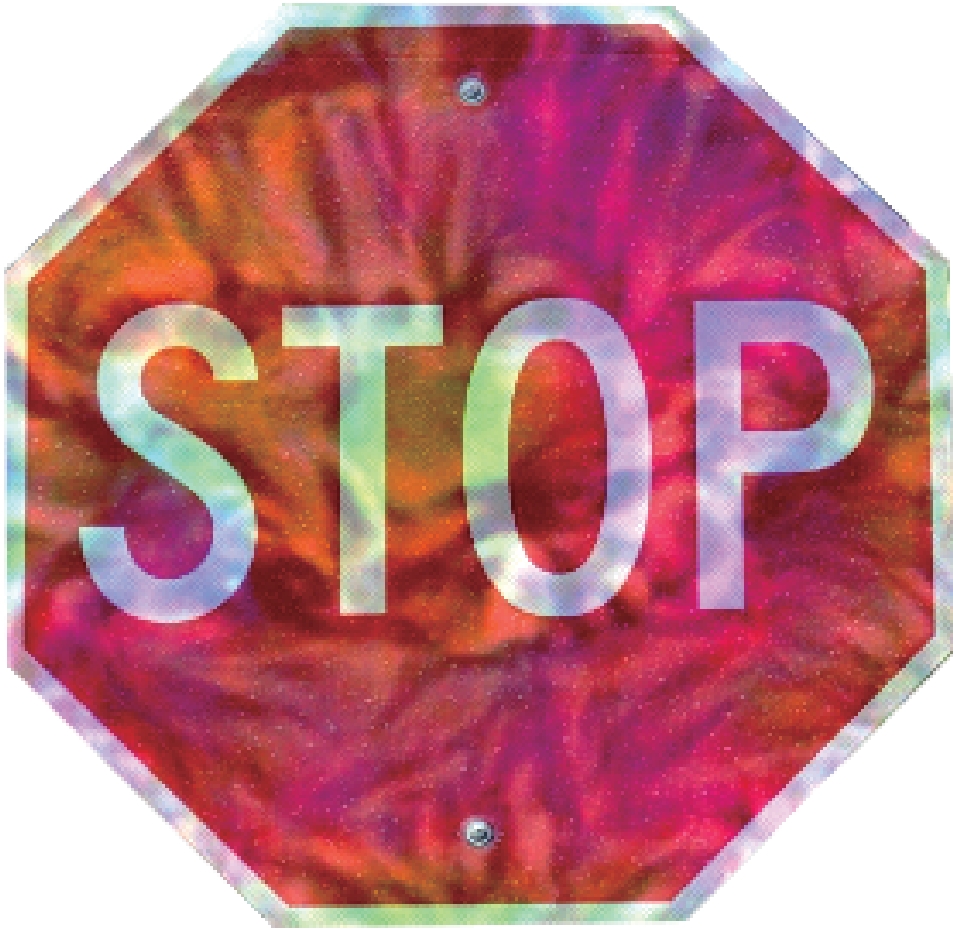}
		\centerline{\footnotesize  Lu \etal~\cite{lu2017adversarial}} 
	\end{minipage}
	\begin{minipage}{.3\linewidth}
		\centering
		\includegraphics[width =1\linewidth]{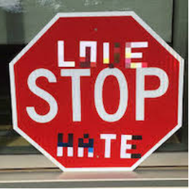}
		\centerline{\footnotesize RP2 \cite{eykholt2018robust}} 
	\end{minipage}
        \begin{minipage}{.18\linewidth}
		\centering
		\includegraphics[width =1\linewidth]{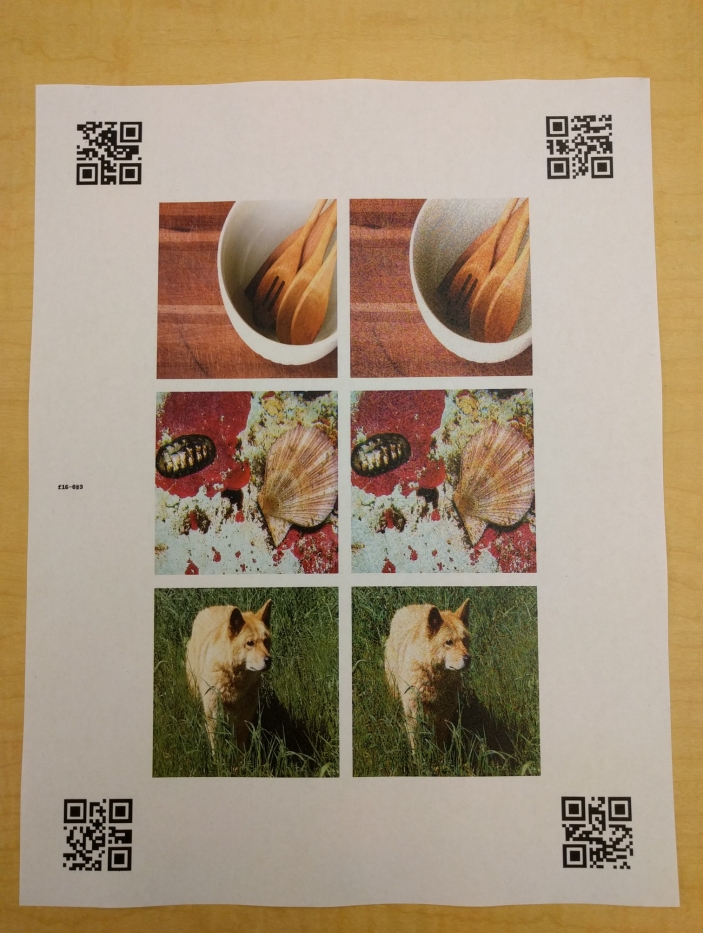}
		\centerline{\footnotesize Kurakin \etal~\cite{ifgsm2018adversarial}} 
	\end{minipage}
        \begin{minipage}{.24\linewidth}
		\centering
		\includegraphics[width =1\linewidth]{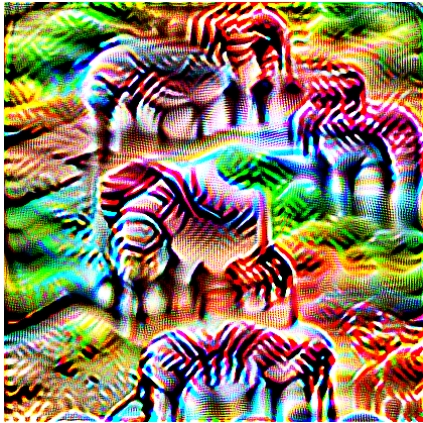}
		\centerline{\footnotesize \quad Lee \etal~\cite{lee2019physical}} 
	\end{minipage}
        \begin{minipage}{.24\linewidth}
		\centering
		\includegraphics[width =1\linewidth]{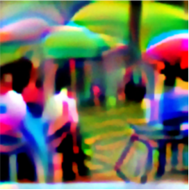}
		\centerline{\footnotesize Thys \etal~\cite{Thys2019fooling}} 
	\end{minipage}
        \begin{minipage}{.24\linewidth}
		\centering
		\includegraphics[width =1\linewidth]{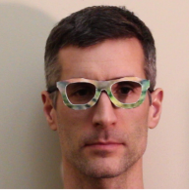}
		\centerline{\footnotesize Sharif \etal~\cite{sharif2019general}} 
	\end{minipage}
        \begin{minipage}{.3\linewidth}
		\centering
		\includegraphics[width =1\linewidth]{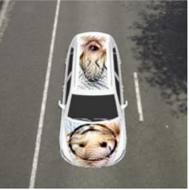}
		\centerline{\footnotesize DAS \cite{DBLP:conf/cvpr/WangLYLTL21}} 
	\end{minipage}
         \begin{minipage}{.3\linewidth}
		\centering
		\includegraphics[width =1\linewidth]{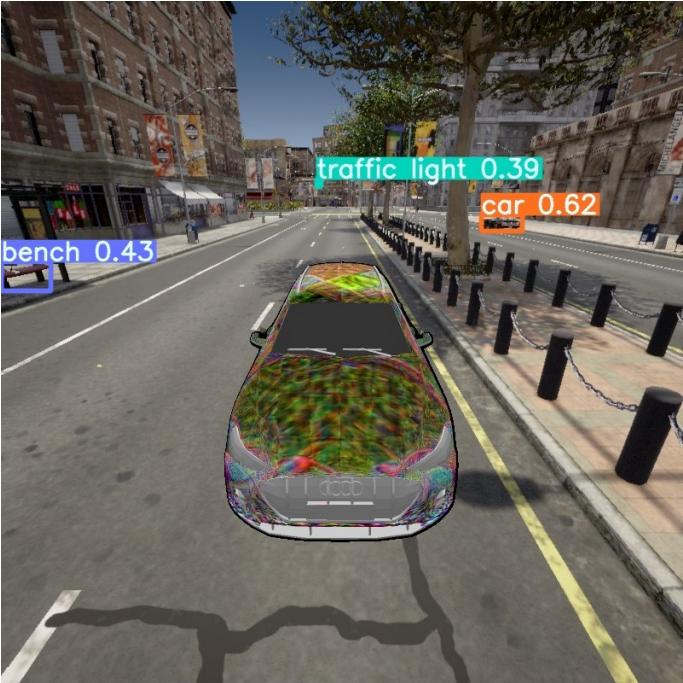}
		\centerline{\footnotesize FCA \cite{wang2022fca}} 
	\end{minipage}
        \begin{minipage}{.3\linewidth}
		\centering
		\includegraphics[width =1\linewidth]{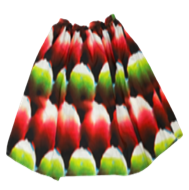}
		\centerline{\footnotesize Hu \etal~\cite{hu2022adversarial}} 
	\end{minipage}

	\caption{Examples of physical adversarial example of touchable attacks.}
	\label{fig:touchable}
\end{figure}
\ding{183} \emph{\textbf{3D attacks}}.
This part discusses the problem of the rendering and distortion simulation of adversarial examples in a 3-dimensional world, including texture on steel bodies, flex bodies, and 3D objects. The first 3D adversarial attack is proposed by Athalye \etal~\cite{athalye2018synthesizing}, for attacking classifiers.\par
To simulate surface materials in 3D space, one way is to simulate them using 2D patches added at specific locations in the image. However, the disadvantage of this approach is that the spatial transformations are quite different from those of a real scene.
Huang \etal~propose the universal physical camouflage (UPC) attacks against object detection models \cite{huang2020universal}. They introduce \textbf{\textit{AttackScenes}} into the physical adversarial camouflage training process and make efforts to effectively handle non-rigid or non-planar objects.
Wang~\etal~\cite{DBLP:conf/cvpr/WangLYLTL21} leverage neural renderer to generate natural adversarial camouflage for vehicles like cars.
Wang~\etal~\cite{wang2022fca} present a workflow to train the full-coverage adversarial camouflage. They use the differentiable neural renderer to render the vehicle and fuse the rendered object with an original image with the mask generated by the semantic segmentation model. 
To further enhance the operability and maintain the adaptability to the physical environment in real-world attack, Zhang~\etal~\cite{zhang2023boosting} perturb only the texture on the top surface of the vehicle and simulate the UAV remote sensing scenario using the backgrounds captured from the CARLA simulator.

Recent research has revealed new insights into the properties of the flex body, in addition to the well-known texture of the steel body. Xu \etal~\cite{xu2020adversarial} first manufactured an adversarial T-shirt by developing a transformer based on Thin Plate Spline (TPS) that could accurately model the effect of deformation. Let $\mathcal{L}_{adv}$ be an attack loss for misdetection, then the optimization process can be formulated as:
\begin{equation}
\begin{split}
     \mathop{min}_\delta \sum_i \mathbb{E}_{t,t_{TPS},v}\mathcal{L}_{adv} + \lambda \mathcal{L}_{tv},
\end{split} 
\label{equ:tshirt}
\end{equation}
where $\mathbb{E}$ denotes environment conditions containing a TPS transformation $t_{TPS} \in \mathcal{T}_{TPS}$, a conventional transformation $t \in \mathcal{T}$ and a Gaussian noise $v$. Considering the difference between flexible and rigid materials, Hu \etal~\cite{hu2022adversarial} utilized the toroidal cropping method to manufacture arbitrary length and expandable adversarial texture.




Furthermore, objects with specific shapes can also become adversarial examples for 3D vision models. 3D printing makes it easier to create shape-specific adversarial examples, which poses a safety risk for the LiDAR-based intelligent sensing system, which is more critical in autonomous driving.
Athalye \etal~\cite{athalye2018synthesizing} first manufactured physical adversarial objects with 3D-printing. 
Based on a differentiable renderer, Xiao \etal~\cite{xiao2019meshadv} computed accurate shading and manufactured adversarial objects with rich shape features but minimal texture variation.
Tu~\etal~\cite{tu2020physically} proposed the first physically realizable adversarial example for LiDAR object detection.
To ensure the realizability of the 3D adversary, they used Laplacian loss to regularize the mesh geometry and applied the rotation and position transformation as prior literature in graphics~\cite{Liu2019SoftRA}.
Miao~\etal~\cite{miao2022Isometric3DAdv} considered the naturalness and invariance of physical transformations of the manufactured adversarial 3D objects.
They used the Gaussian curvature as a metric to measure and optimize the naturalness, and approximated the worst condition in the physical-to-digital transformation (MaxOT) for robustness. 

Recently, Liu and Guo \etal~\cite{liu2023x} innovatively propose the first physical adversarial attacks against the X-ray prohibited item detection task and tackle the key issues in the feasibility of physical-world adversarial attacks in the visually-constrained X-ray imaging scenario. 
In detail, they design the $\mathcal{X}$-adv to generate the physically realizable 3D objects with adversarial shapes aiming to manufacture, which enable the attacks to remain effective when facing color/texture fading. Similarly, Lou \etal~\cite{lou2024hide} generate adversarial objects against 3D points clouds.

\subsubsection{Untouchable Attacks}
The untouchable attacks consist of lighting attacks. The generated PAEs of these attacks could not be touched in the air.


The fabrication of adversarial attack examples is not limited to physical objects; effective adversarial attacks can be generated using light as well.
The key in modeling lightning attacks is to build constraints to make the perturbation realizable using optical attack devices and to find controllable parameters in the physical world.
The representation characteristics of lighting attacks are stealthy and controllable, which means that the attacker can perform attacks or not by turning on/off the light source, raising a greater challenge to the security of deep models in real scenarios. Until now, a line of research has been developed to leverage the light to perform physical attacks.

Nichols \etal~\cite{nichols2018projecting} first demonstrated that using one red point light source merely project toward the toy car can effectively mislead the DNNs, where only the location of light on the car is optimized by the differential evolution algorithm like the one-pixel attack. Unlike \cite{nichols2018projecting} emitted only one point light, Man \etal~\cite{man2019poster} projected the light by a projector to cover the whole area of the target object, where the projection color is exhaustively optimized. Compared to \cite{nichols2018projecting}, \cite{man2019poster} introduced a large modification on the target object, making it conspicuous to human observers. 
Instead of constructing the light, Guo \etal~\cite{guo2020watch} exploited the blurring effect caused by object motion to perform attacks against the DNNs, where the blurring effect is produced by the transform parameters for the foreground and background.

Differ from \cite{nichols2018projecting,man2019poster} that focused on emitting the light toward the target object, Kim \etal~\cite{kim2021light} considered the imaging process and developed a physical attack by placing a spatial light modulator (SLM) in the front of the photographic system, changing the phase of the light in Fourier domain. 

\textit{Sayles et.al.}~\cite{sayles2021invisible} presented adversarial attack using programmable LED light.
According to the \textit{Rolling Shutter Effect}~\cite{Liang2008AnalysisAC} in digital rolling shutter camera, the pixels with different vertical coordinate has different exposure time, which enables the authors to perturb the captured image with the high-frequency ambient source with stripe patterns.
The perturbation generation process can be formulated as follows within the context of $\mathcal{T}$ modeling environment conditions $\mathbb{C}$, which also correlates to $\mathcal{R}(\cdot)$ mentioned previously, during the re-sampling process. Let $x_{amb}$ represent the image captured under ambient light conditions, $x_{sig}$ denote the image taken under the influence of fully illuminated attacker-controlled lighting, and $g(y+\delta)$ indicate the average impact of the signal on row $y$:
\begin{equation}
\begin{split}
     x_{adv}^p = \mathcal{T}(x_{amb})+\mathcal{T}(x_{sig})\cdot g(y+\delta).
\end{split} 
\label{equ:invisible}
\end{equation}

Unlike using the light to modify the RGB image, Zhu \etal~\cite{zhu2021fooling} leveraged the board placed with optimized bulbs to modify the infrared imaging result and mislead the object detector to output the wrong result. In their attack, the position and shining of the bulb are optimized, where the latter is modulated by controlling the connection circus.

However, due to the vulnerability of DNN, a more simple lightning model can be constructed. 
\textit{AdvLB}~\cite{Duan2021AdversarialLB} shows that DNNs are easily fooled by only using a laser beam, and shows effectiveness in both the digital and physical world.
The authors modeled the laser beam as an additional rendering layer on top of the image, and the laser beam is formed with only a few parameters, consisting of (1) wavelength, (2) width, (3) intensity, (4) angle, and intercept, that determines the laser beam in the image coordinate.
Then these parameters are optimized by a simple randomized greedy search. 
Pony \etal~\cite{pony2021over} proposed a flickering attack method for video recognition. 
The flickering source is an RGB LED that controls the global ambient light, which can be simply modeled by parameter $\delta \in \mathbb{R}^{T\times3}$ where $T$ is the length of the video.
Gnanasambandam \etal~\cite{gnanasambandam2021optical} modeled the project-to-camera process by studying the mechanism of the radiometric response of the projector and the spectral response of the scene, which is introduced into the optimization of the adversarial attack. Based on the projector-to-camera model, the author provided a theoretical guarantee for their attack algorithm, making it interpretable and transparent to the user. 
Concurrent to \cite{gnanasambandam2021optical}, Lovisotto \etal~\cite{lovisotto2021slap} elaborately manufactured the projector-based adversarial attack. The author simulated the projection process by training a neural network that takes the color that the projector can be represented and camera noise as input and outputs the actual projected color in the real world. 

Other works focus on attacking real applications and specific tasks. 
Speth \etal~\cite{speth2022digital} presents the first examples of digital and physical attacks on remote photoplethysmography (rPPG), a technique for estimating blood volume changes from reflected light without contact sensors. The authors show how to add imperceptible noise to the input videos or use visible spectrum LEDs to fool rPPG systems and generate a fake blood volume pulse. To model the reliable adversarial attack using RGB ambient light, they constrain the gradient to be spatially consistent.
Huang \etal~\cite{huang2022spaa} developed a stealthy projector-based adversarial attack by formulating the projector-to-capture problem involving in physical attack as an end-to-end differentiable process, significantly alleviating the distortion caused by camera-captured adversarial projector's light. 
Similar to \cite{kim2021light}, Hu \etal~\cite{hu2022advlen} explored the influence of zoom out/in of the camera sensor len on the follow-up process of DNN inference, where the optimal coefficient of zoom is optimized to guarantee the obtained image can deceive the DNNs. Like \cite{man2019poster}, Hu \etal~\cite{hu2022advfilm} developed a physical attack method by placing the color transparency sheet in front of the camera sensor to engender a film-style color over the object. 
Li \etal~\cite{li2023physical} generated adversarial illumination against 3D face recognition systems via attacking the structured light profilometry with the Lambertian rendering model. 

Recently, some natural phenomena (e.g., shadow \cite{zhong2022shadows} and reflected light \cite{wang2023rfla}) have been exploited to perform physical adversarial attacks. Zhang \etal~\cite{zhong2022shadows} modeled the shadow that naturally exists in the real world to perform a stealthy attack, where the position and geometry (determined by points) of the shadow on the target object are optimized by a particle swarm algorithm (PSO). Follow-up \cite{zhong2022shadows}, Wang \etal~\cite{wang2023rfla} pointed out the shortcomings of the previous optical-based method that was invalid in the daylight and proposed to utilize the reflected light to perform a physical attack, which is insusceptible to sunlight. Moreover, it can be deployed in the evening using a flashlight source. 
Further, Schmalfuss~\etal~\cite{schmalfuss2023distracting} showed that particle rendering can also be used to model natural phenomena, such as snow and rain, and adversarially optimized particles can effectively attack motion estimation models.

\begin{figure}[t]
	\centering
	\begin{minipage}{.26\linewidth}
		\centering
		\includegraphics[width =1\linewidth]{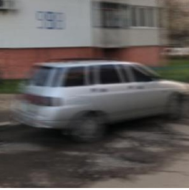}
		\centerline{\footnotesize  Guo \etal~\cite{guo2020watch}} 
	\end{minipage}
	\begin{minipage}{.26\linewidth}
		\centering
		\includegraphics[width =1\linewidth]{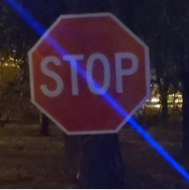}
		\centerline{\footnotesize Duan \etal~\cite{Duan2021AdversarialLB}} 
	\end{minipage}
	\begin{minipage}{.38\linewidth}
		\centering
		\includegraphics[width =1\linewidth]{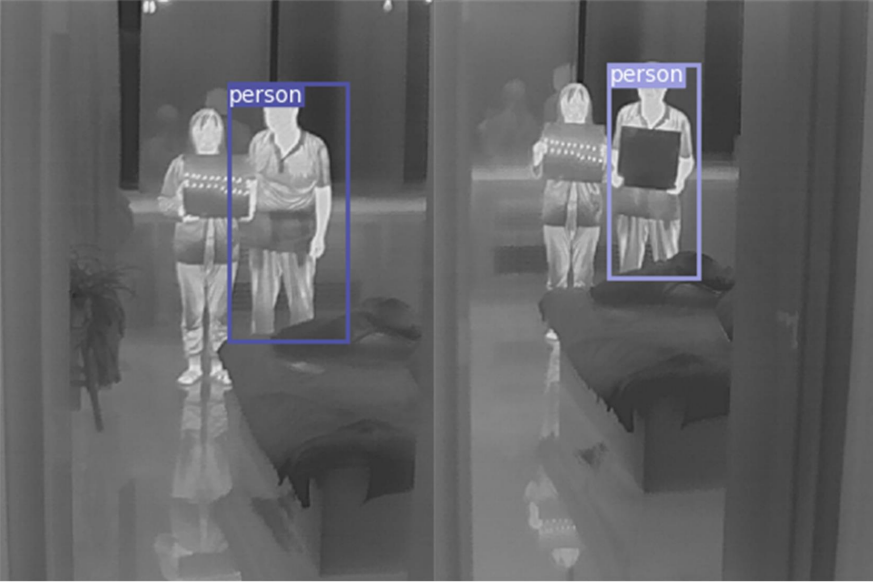}
		\centerline{\footnotesize Zhu \etal~\cite{zhu2021fooling}} 
	\end{minipage}
        \begin{minipage}{.3\linewidth}
		\centering
		\includegraphics[width =1\linewidth]{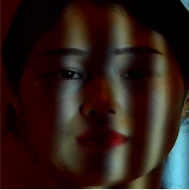}
		\centerline{\footnotesize Li \etal~\cite{li2023physical}} 
	\end{minipage}
        \begin{minipage}{.3\linewidth}
		\centering
		\includegraphics[width =1\linewidth]{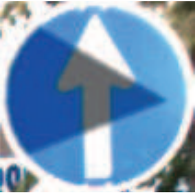}
		\centerline{\footnotesize Zhong \etal~\cite{zhong2022shadows}} 
	\end{minipage}
        \begin{minipage}{.3\linewidth}
		\centering
		\includegraphics[width =1\linewidth]{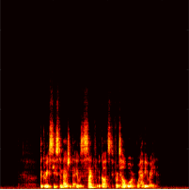}
		\centerline{\footnotesize Li \etal~\cite{li2020practical}} 
	\end{minipage}
	\caption{Examples of physical adversarial example of untouchable attacks.}
	\label{fig:untouchable}
\end{figure}

\subsection{The re-sampling Process Oriented PAEs}
\begin{figure}
    \centering
    \includegraphics[width=0.99\linewidth]{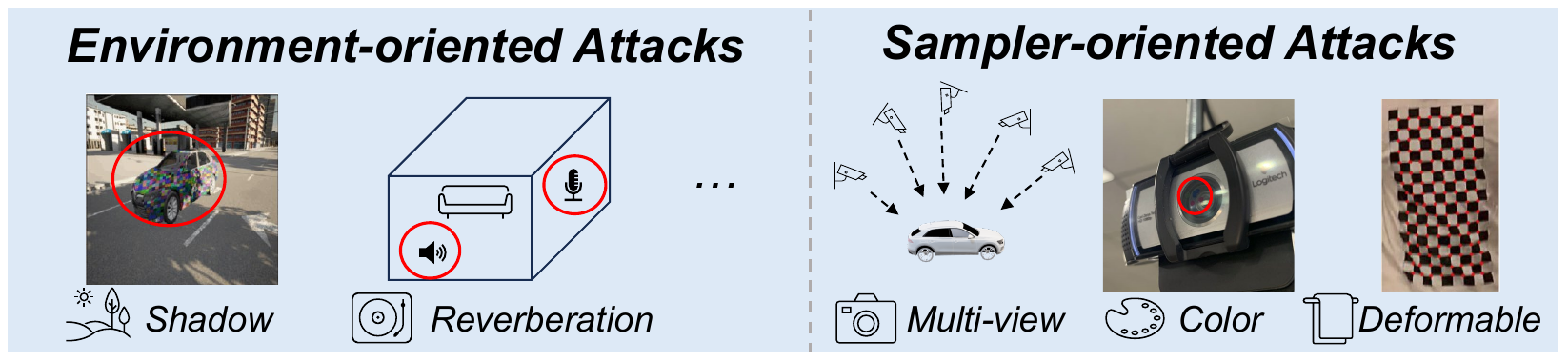}
    \caption{The overview of the re-sampling process oriented PAEs.}
    \label{fig:resample}
\end{figure}

After finishing manufacturing, the physical adversarial examples will take effect by being re-sampled and input into the deployed deep models in real artificial systems. And during this process, some of the key information correlated to the adversarial characteristics inside the PAEs might be affected and cause certain attacking ability degeneration due to the imperfect re-sampling, which could be also called physical-digital domain shifts. More precisely, this kind of physical-digital shift consists of 2 types as shown in Figure \ref{fig:resample}, \ie, the environment-caused and sampler-caused, therefore motivating us to categorize the re-sampling process-oriented PAEs into environment-oriented attacks and sampler-oriented attacks for better understanding.

\subsubsection{Environment-oriented Attacks}
During the re-sampling process in real scenarios, the existing noises in the environment will show significant effectiveness to the adversarial examples. For instance, the lighting and weather could affect the imaging of the adversarial examples in the computer vision area, and the environmental noises from devices and wild might cause the distortion of the adversarial audio examples. To overcome these challenges, the researchers make efforts the environment-oriented attacks to generate adversarial examples under the consideration of the environmental shifts that might cause attacking ability degeneration. 

\begin{figure}[t]
	\centering
	\begin{minipage}{.3\linewidth}
		\centering
		\includegraphics[width =1\linewidth]{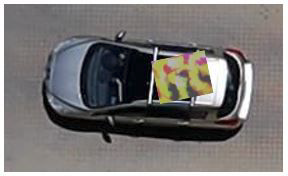}
		\centerline{\footnotesize  Du \etal~\cite{Du2021PhysicalAA}} 
	\end{minipage}
	\begin{minipage}{.33\linewidth}
		\centering
		\includegraphics[width =1\linewidth]{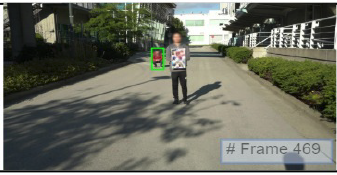}
		\centerline{\footnotesize Ding \etal~\cite{ding2021towards}} 
	\end{minipage}
	\begin{minipage}{.3\linewidth}
		\centering
		\includegraphics[width =1\linewidth]{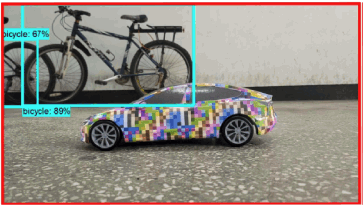}
		\centerline{\footnotesize DTA \cite{suryanto2022dta}} 
	\end{minipage}
	\caption{Examples of PAE of environment-oriented attacks.}
	\label{fig:environment}
\end{figure}

The physical attack performance is significantly impacted by environmental factors, such as light and weather, which motivates the researcher to take these factors into account during the optimization of PAEs. Du \etal~\cite{Du2021PhysicalAA} proposed the physical adversarial attack for aerial imagery object detector, avoiding remote sensing reconnaissance. They design the tools for simulating re-sampling differences caused by atmospheric factors, including lightning, weather, and seasons. Finally, they optimize the adversarial patch by minimizing the following loss function.
\begin{equation}
\mathcal{L} = \mathbb{E}_{t\in \mathcal{T}}[\max(\mathcal{F}^b(t(x^d_{adv})))] + \lambda_1 \mathcal{L}_{nps}(\delta) + \lambda_2 \mathcal{L}_{tv}(\delta),
\end{equation}
where the first term is the adversarial loss to suppress the maximum prediction objectness score over the transformation distribution $\mathcal{T}$, the second term ensures the optimized color is printable, and the last term is used to ensure the naturalness of the adversarial patch.
Ding \etal~\cite{ding2021towards} took the potential transformation (e.g., brightness, contrast, color) that may occur in the real world to enhance the physical attack of the universal adversarial patch against siamese-based visual-based object tracker.
Suryanto \etal~\cite{suryanto2022dta} revealed that current camouflage-based attacks \cite{DBLP:conf/cvpr/WangLYLTL21,wang2022fca} ignore the shadow that may impact the attack performance in the real world. Thus, the author devised a differentiable transformation network (DTN) to learn potential physical transformations (e.g., shadow). 
The adversarial LED light attack \cite{sayles2021invisible} also concerns the environment inside the attacking scenario.
During the perturbation generation process, they propose the function $\mathcal{T}$,  which can be regarded as the $\mathcal{R}(\cdot)$ in our definition, to model environment conditions (including viewpoint and lighting changes) and preserve the attacking ability and cross the digital-physical domain.

Unlike images, speech has its own physical characteristics, such as the reverberation caused by the speech crashing other objects in the environment, significantly impacting speech physical attack performance. To alleviate the potential distortion caused by the environment, a line of works \cite{qin2019imperceptible,li2019adversarial,yakura2019robust,li2020practical,xie2020real,zhang2021attack}  has adopted the room impulse response (RIP) to mimic distortion caused by the process of the speech being played and recorded, which can be expressed as:

\begin{equation}
x_{adv}^d(t): r(t) = y^p_{adv}(t) * x_{adv}^d(-t),
\end{equation}
where the $x_{adv}^d(t)$ is the audio clip, and the $y^p_{adv}(t)$ is the corresponding estimated recorded audio clip, $*$ denotes the convolution operation. Then, RIP $r(t)$ incorporates the generation of $\delta$ by a transform $T(x)=x * r$, which reduces the impact of distortion brought by hardware and physical signal patch, significantly improving speech physical attack robustness. For example, Yakura \etal~\cite{yakura2019robust} incorporated an impulse response and white Gaussian noise to enhance the AEs' robustness towards reverberations and background noise. Xie \etal~\cite{xie2020real} built the robust universal perturbation by estimating the RIR with an acoustic room simulator that constructed over 100 locations of the loudspeaker,  which makes them work on various simulator scenarios and generalize well into the real scenario. Recently, Zhang \etal~\cite{zhang2021attack} proposed two approaches to enhance the robustness of physical attacks: the first is to adopt the SineSweep method \cite{stan2002comparison} to estimate the RIP in a room to model the distortion from speaker playing to microphone recording. The second is that the attacker plays the adversarial perturbation as a separate source when the adversary is speaking, which can effectively evade the replay detection module.

In addition to the environment of everyday life, the construction of the confrontation samples was further extended to the industrial environment.
Chen \etal~\cite{chen2022advICS} proposed the adversarial attack for the instruction detection model in industrial control systems (ICS). To achieve successful attacks, they manipulated the network traffic flow with PAEs optimized in the restricted space defined by the ICS.
Regarding machine learning models for locating and detecting false data injection attacks (FDIA) in smart grids, Tian\etal~\cite{Tian2024} proposed multi-label adversarial example attacks to achieve attacks against these models. 
The Direct-current (DC) state estimation model is used as the environment simulator, and the adversarial attack based on it poses a new security threat to smart grids.

\subsubsection{Sampler-oriented Attacks}
Besides the environmental noises (\ie, the kind of objective effectiveness), there also exists some sampler correlated wastage of adversarial information inside the adversarial examples. The case in point is sampling angles in computer vision tasks, when taking photos from different perspectives, the sampled instances might show slight differences in shape and color, \eg, affine-transformation-like difference, and overexpose. Similarly, in audio tasks, the commonly existing differences of various recorders might hurt the attacking ability of adversarial examples.
\begin{figure}[t]
	\centering
	\begin{minipage}{.24\linewidth}
		\centering
		\includegraphics[width =1\linewidth]{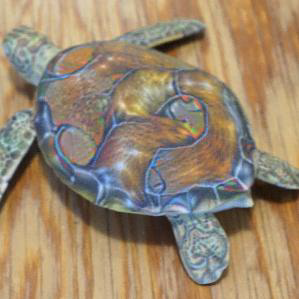}
		\centerline{\footnotesize  EoT\cite{athalye2018synthesizing}} 
	\end{minipage}
	\begin{minipage}{.24\linewidth}
		\centering
		\includegraphics[width =1\linewidth]{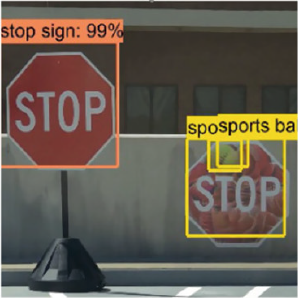}
		\centerline{\footnotesize Shapeshifter \cite{chen2018shapeshifter}} 
	\end{minipage}
	\begin{minipage}{.24\linewidth}
		\centering
		\includegraphics[width =1\linewidth]{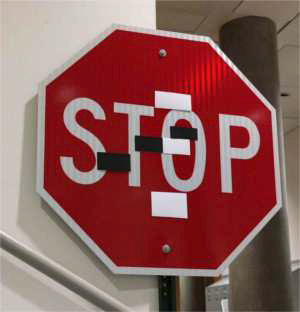}
		\centerline{\footnotesize $RP_2$ \cite{eykholt2018physical}} 
	\end{minipage}
	\begin{minipage}{.24\linewidth}
		\centering
		\includegraphics[width =1\linewidth]{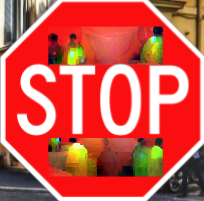}
		\centerline{\footnotesize Eykholt \etal~\cite{eykholt2018robust}} 
	\end{minipage}	
	\begin{minipage}{.33\linewidth}
		\centering
		\includegraphics[width =1\linewidth]{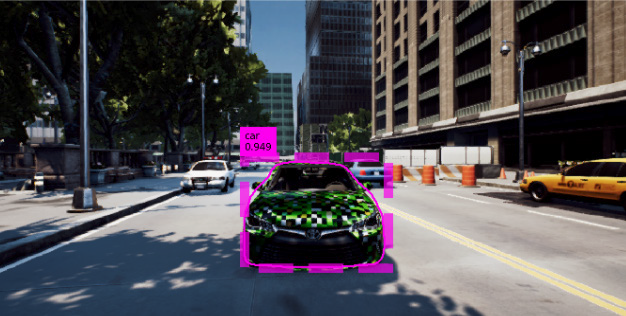}
		\centerline{\footnotesize CAMOU \cite{zhang2018camou}} 
	\end{minipage}
	\begin{minipage}{.31\linewidth}
		\centering
		\includegraphics[width =1\linewidth]{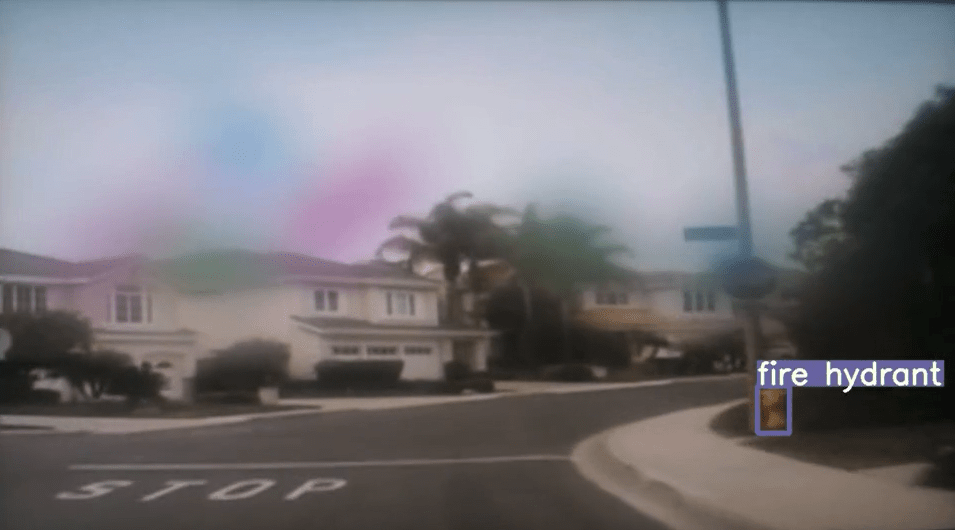}
		\centerline{\footnotesize Translucent Patch \cite{zolfi2021translucent}} 
	\end{minipage}
	\begin{minipage}{.3\linewidth}
		\centering
		\includegraphics[width =1\linewidth]{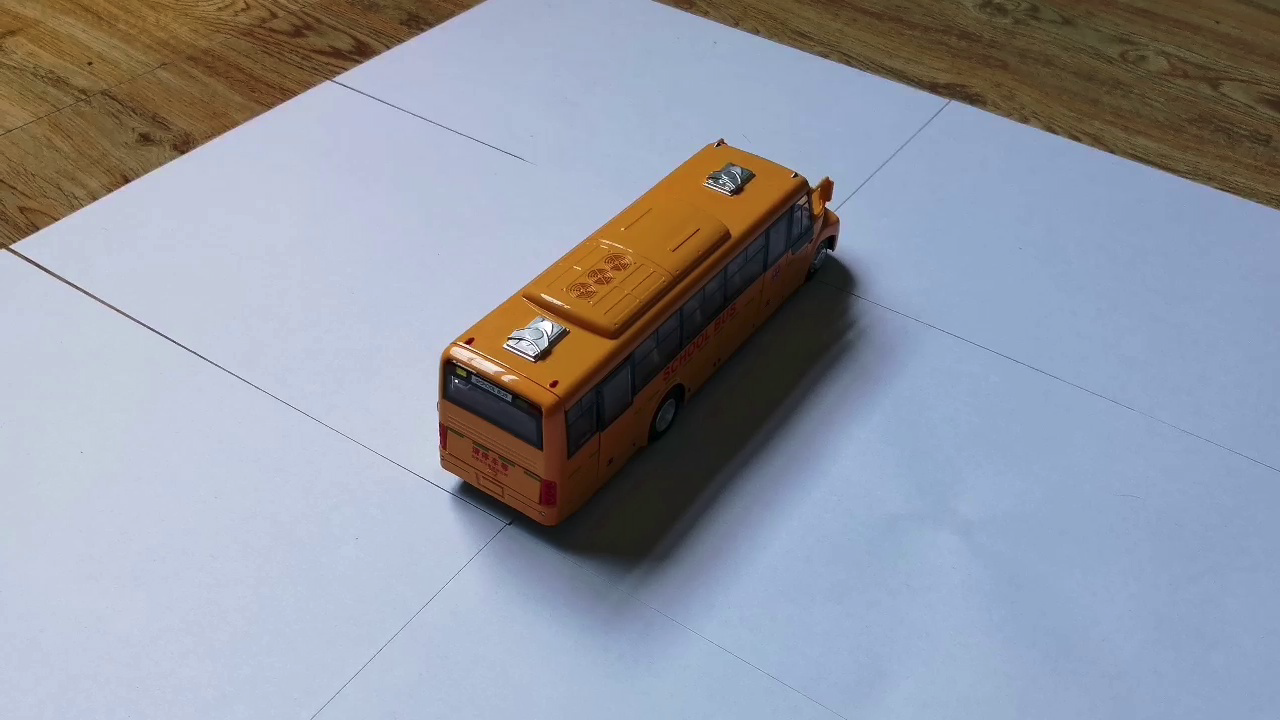}
		\centerline{\footnotesize ViewFool \cite{dong2022viewfool}} 
	\end{minipage}	
	\begin{minipage}{.25\linewidth}
		\centering
		\includegraphics[width =1\linewidth]{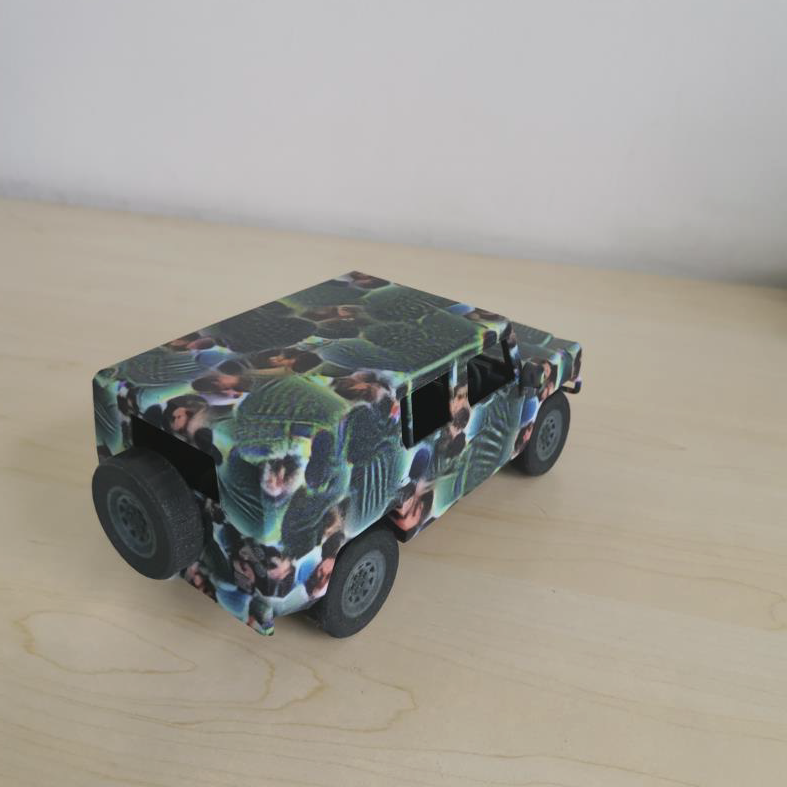}
		\centerline{\footnotesize CAC \cite{duan2022learning}} 
	\end{minipage}
	\begin{minipage}{.25\linewidth}
		\centering
		\includegraphics[width =1\linewidth]{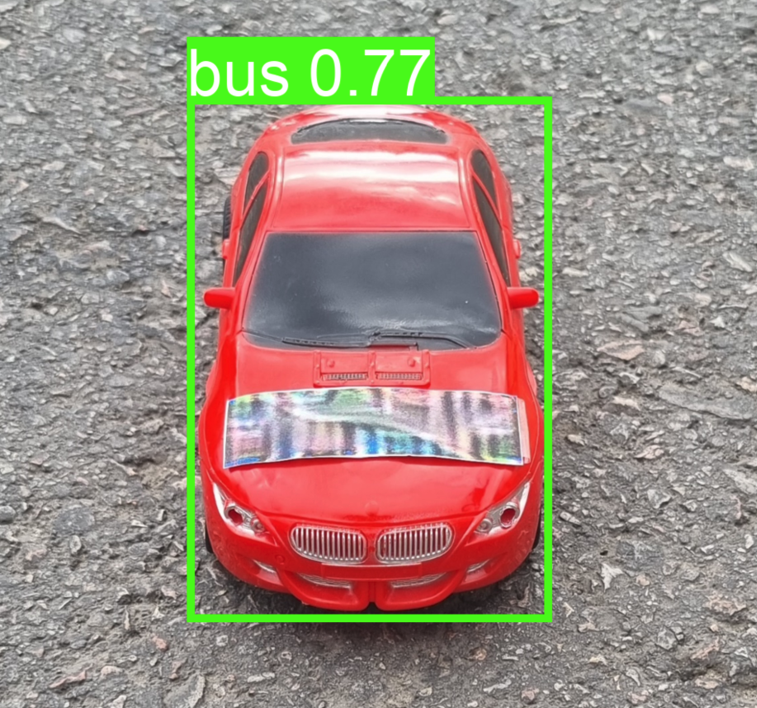}
		\centerline{\footnotesize Shapira \etal\cite{shapira2022attacking}} 
	\end{minipage}	
	\begin{minipage}{.25\linewidth}
		\centering
		\includegraphics[width =1\linewidth]{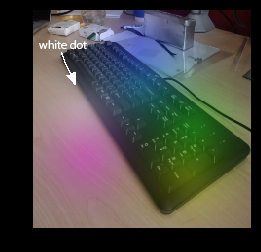}
		\centerline{\footnotesize Li \etal~\cite{li2019sticker}} 
	\end{minipage}
	\begin{minipage}{.21\linewidth}
		\centering
		\includegraphics[width =1\linewidth]{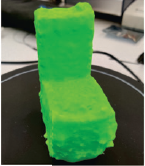}
		\centerline{\footnotesize Tsai \etal~\cite{tsai2020robust}} 
	\end{minipage}
	\caption{Examples of PAE of sampler-oriented attacks.}
	\label{fig:sampler}
\end{figure}

To confront the view perspective change in the physical world, Athalye \etal~\cite{athalye2018synthesizing}  formulated the potential physical transformations (e.g., rotation, scale, resize) as a uniform formal that is the expectation over transformation (EoT), which is mathematically denoted as follows.
\begin{equation}
\delta = \mathbb{E}_{t \sim \mathcal{T}}[d(t(x^d_{adv}), t(x))].
\end{equation}
The above formula is designed to alleviate the data domain gap caused by transformation, enhancing the robustness of the adversarial texture. In addition, they fabricated 3D physical-world adversarial objects for image classification models, wherein the physical attack robustness of the object's adversarial texture is enhanced after training with the chosen transformation distributions. Chen \etal~\cite{chen2018shapeshifter} developed a ShapeShifter attack against Faster RCNN, which was enhanced by EoT for robust physical attack. Follow \cite{athalye2018synthesizing}, Chen \etal~\cite{chen2018shapeshifter} developed a ShapeShifter attack against Faster RCNN, which was also enhanced by EoT for robustness. 

Different from utilized two-dimensional transformation \cite{athalye2018synthesizing,chen2018shapeshifter}, some researchers attempt to use other methods to boost the physical attack to confront the perspective difference. Eykholt \etal~first in \cite{eykholt2018robust} sampled a large-scale multi-view physical dataset, which is used to train the optimal white-black bock perturbation to attack the traffic sign recognition model. Specifically, they utilized the mask to constrain the perturbation to be located in the traffic sign area, and the position of the perturbation is optimized by imposing the $L_1$-norm. The above optimization can be expressed as 
\begin{equation}
\arg \min_\delta \lambda ||\delta||_p + \mathcal{L}_{nps}(\delta) + \mathbb{E}_{x \sim X^{V}}\mathcal{L}(\mathcal{F}(x+t(\delta)), y),
\end{equation}
where the first term is used to bound the norm of $\delta$ for the patch's imperceptible, the third term takes into account the transformation inside in $x$ and applies the same transformation on $\delta$ and the $X^{V}$ includes the digital and physical collected training dataset.

Follow-up \cite{eykholt2018physical}, they devised a synthesis method in 3D space to mimic the rotation and position transformation of the object in the physical world; the adversarial sticker optimized with that method can successfully make the traffic sign ``disappear" under the view of the object detector. Similar to \cite{eykholt2018robust}, Wang \etal~\cite{wang2019advpattern} first sampled enormous images of the person from different viewpoints of the camera, while the difference is that they sampled from the simulator environment rather than the real world. Then, they optimized a universal patch with the devised loss function to realize the goal of hiding one person from the re-identification model. Moreover, the author empirically found that the NPS loss is not helpful to physical attack performance, as it is hard to balance with other objectives. Concurrent to \cite{wang2019advpattern}, Zhang~\etal~\cite{zhang2018camou} learned the adversarial camouflage for a vehicle with the simulated environment provided by open-source simulator AirSim. To address the non-differentiable of the simulator, they devised a clone network to mimic the whole process consisting of wrapping the adversarial camouflage over the vehicle, then rendering it into the 2D image with a physical renderer, and finally getting the detection result. At the same time, they perform a white-box attack against the clone network to update the adversarial camouflage. Concurrently, Zeng \etal~\cite{zeng2019adversarial} paid attention to 3D physical properties instead of 2D images, and the author proposed to simulate physical transformations (e.g., rotation in 3D space) augmented DNNs with a rendering module. Duan \etal~\cite{duan2022learning} utilized the physical render technique proposed in \cite{athalye2018synthesizing} to optimize the adversarial UV texture for the vehicle. The adversarial UV texture is wrapped over the vehicle by changing the camera position and rendered into multi-view images. Thus, the adversarial UV texture is trained to optimize the following object function
\begin{equation}
\arg \min_{\delta} \mathbb{E}_{x\sim X, e \sim \textbf{E}}[\frac{1}{n}\sum_{p_i\in P} \mathcal{L}(\mathcal{F}(x^d_{adv}, p_i), y)],
\end{equation}
where \textbf{E} denotes the environment condition determined by the physical render, such as different viewpoints and distances (here the \textbf{E} is different from the real environment $\mathcal{C}$); $P$ indicates the output proposals of each image respective to the two-stage detector (e.g., Faster RCNN). In such a way, they can mimic the perspective changing in the physical world as possible. Recently, Shapira \etal~\cite{shapira2022attacking} tailored the projection function to place the adversarial patch on the target, where the size and perspective of target objects vary due to the car moving and road scenario in traffic surveillance. In such a way, the obtained patch remains effective when it is placed anywhere on the car. To solve the problem that far points after surface reconstruction are easily removed by defense mechanisms, Tsai \etal~\cite{tsai2020robust} the authors utilized kNN distance as well as perturbation clipping and projection to optimize a physically deployable adversarial 3D objects against point clouds. Recently, Dong \etal~\cite{dong2022viewfool} demonstrated that there exist adversarial viewpoints, where images captured under such viewpoints are hard to recognize for DNN models. They leveraged the Neural  Radiance Fields (NeRF) technique to find the adversarial viewpoints. Specifically, they find the adversarial viewpoints by solving the following problem
\begin{equation}
\max_{p(v)} \left\{\mathbb{E}_{p(v)} [\mathcal{L}(\mathcal{F}(\mathcal{G}(v)), y)] + \lambda \cdot \mathcal{H}(p(v)) \right\},
\end{equation}
where $p(v)$ denotes the adversarial viewpoints distribution, $\mathcal{G}(v)$ is the render function of NeRF, which renders an image with the input viewpoints; $ \mathcal{H}(p(v))=\mathbb{E}_{p(v)}[-\log(p(v))]$ is the entropy of the distribution of $p(v)$. 

In addition to the view perspective, the discrepancy property among different objects may also impact the attack performance, such as non-rigid objects being easily deformable, whereas the accessory (e.g., the attached adversarial patch) would be deformable simultaneously. To alleviate the influence of deformable, Xu \etal~\cite{xu2020adversarial} took the Think Plate Spline (TPS) \cite{Bookstein89TPS} method into account when optimizing the wearable adversarial patch to model the topological transformation from texture to cloth caused by body movement. Specifically, they construct the adversarial examples as following
\begin{equation}
x^d_{adv} = t_{env}(A+t(B-C + t_{color}(M_{c,i}\circ t_{TPS}(\delta + \mu v))),
\end{equation}
where $t_{env} \in \mathcal{T}$ indicates the environmental brightness transform, $t_{color}$ is a regression model that learns the color covert between the digital image and its printed counterpart, $t_{TPS}$ denotes the TPS transform; $A$ is the background region expect the person, $B$ is the person-bounded region, and $C$ is the cloth region of the person, $v\in \mathcal{N}(0,1)$ to improve the diversity of perturbation.

For simulating the printing of texture on clothes, Hu \etal~\cite{hu2022adversarial} provides the toroidal-cropping technique that simulates the transformation using sampling.
Following this work, Hu~\etal~\cite{hu2023physically} further integrated the 3D TPS as topological transformation, projection as geological transformation and the UV coordinates of the cloth from a 3D graphic model to generate physically realizable adversarial camouflage clothing.
Furthermore, in order to fool a wider range of sensors, the material of clothing is not limited to fabric.
Zhu \etal~\cite{zhu2022infrared} uses the QR code-like or checkerboard-like adversarial cloth with heat-insulating material to fool the infrared detectors.

Unlike the view perspective and shape deformable, the color discrepancy is the more common factor that impacts the attack performance. In physical attacks, color distortion may occur in the camera imaging process. To eliminate such influence, Li \etal~\cite{li2019sticker} attacked the re-sampling process directly by placing a carefully-crafted sticker over the camera. To construct unconscious adversarial perturbations, the authors first printed a small dot on transparent paper and collected a pair of benign and adversarial images. Then, the authors maximized the structural similarity (SSIM) to make the images closer. Specifically, they find the patch by solving the following optimization problem
\begin{equation}
\max_{\pi \in \prod} \mathbb{E}_{x\in D(x|y^*)}[\mathcal{L}(\mathcal{F}(\pi(x)), y^*) - \mathcal{F}(\pi(x)), y_t)],
\end{equation}
where $\pi \in \prod$ is an applied function used to find some locations and colors of the adversarial patch. The above problem is devised to find the universal adversarial patch by pushing the adversarial examples close to the target class and away from its original class.

Similar to \cite{li2019sticker}, ~\cite{zolfi2021translucent} proposes a translucent patch that is placed on the camera lens to fool object detectors. The patches are described as 2D images with four channels, including alpha and RGB. The patch is modeled by the alpha blending process in the digital world. Manufacturing distortion and constraints, including smoothness requirement, upper bound of alpha channel, and inaccurate placement, are taken into account. The author further demonstrates the transferability of the adversarial patch across different object detectors.
To construct the imperceptible speech physical attack, Szurley \etal~\cite{szurley2019perceptual} constrained perturbation to beneath the threshold of auditory masking, which is realized by Discrete Fourier Transform (DFT). Concurrent to \cite{szurley2019perceptual}, Qin \etal~\cite{qin2019imperceptible} estimated the masking threshold by using a psychoacoustic model. The approach of psychoacoustic effect proposed by \cite{qin2019imperceptible} has been widely adopted \cite{li2019adversarial}. 

Furthermore, it is necessary to conduct research on re-sample processes with real-world application system taking into account to uncover the real-world threats posed by adversarial examples.
Wang~\etal~\cite{wang2023does} proposed SysAdv, a system-driven attack design in the autonomous driving context, which effectively achieves system-level effects, such as traffic rule violations, significantly improving the attack success rate compared to existing designs that only evaluate the targeted AI component level.

\subsection{Other PAE Topics}
Building upon the manufacturing process-oriented PAEs and re-sampling process-oriented PAEs, there are also some physical adversarial attacks that take into account the critical attributes of PAEs to better perform in real scenarios. For example, as shown in Figure~\ref{fig:natural_examples}, many researchers are concerned about the naturalness of the PAEs, therefore aiming at designing PAEs with a more stealthy appearance in CV tasks.
Similarly, the transferability and the generalization of PAEs are also focused due to the opening of the real world, \ie, the physical world examples are required to be effective against diverse deployed models and various instances.
In this section, we introduce some physical adversarial attacks that are not, or not only, designed for the particular workflow of physical adversarial attacks (although they may involve these particular workflows).
Note that some of the critical attributes of PAEs are bound with the manufacturing process and re-sampling process, \eg, the complexity is tightly correlated to the manufacturing process, the robustness is correlated to the re-sampling process, and the aggressiveness is the indispensable attribute of any adversarial attacks. Thus, we do not discuss these attributes in this section.

\subsubsection{The Natural Physical Adversarial Attacks}

\begin{figure}[t]
	\centering
	\begin{minipage}{.49\linewidth}
		\centerline{\footnotesize SemanticAdv~\cite{qiu2020semanticadv}} 
		\centering
		\includegraphics[width =1\linewidth]{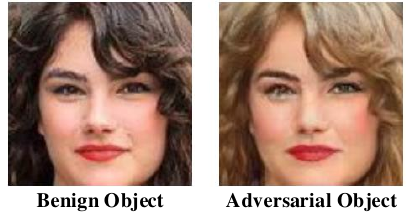}
	\end{minipage}
	\begin{minipage}{.49\linewidth}
		\centerline{\footnotesize  AdvCam~\cite{duan2020adversarial}} 
		\centering
		\includegraphics[width =1\linewidth]{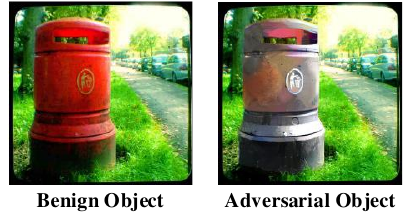}
	\end{minipage}
        \begin{minipage}{.49\linewidth}
		\centerline{\footnotesize Cheng~\etal~\cite{cheng2022physical}} 
		\centering
		\includegraphics[width =1\linewidth]{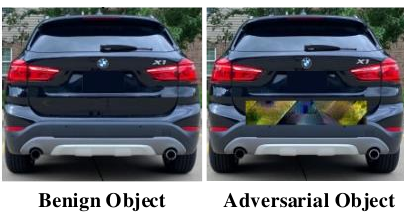}
	\end{minipage}
    \begin{minipage}{.49\linewidth}
    \centerline{\footnotesize PS-GAN\cite{Liu2019PerceptualSensitiveGF}} 
    \centering
    \includegraphics[width =1\linewidth]{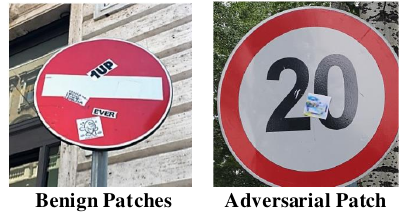}
	\end{minipage}
	\begin{minipage}{.49\linewidth}
		\centerline{\footnotesize LAP~\cite{tan2021legitimate}} 
		\centering
		\includegraphics[width =1\linewidth]{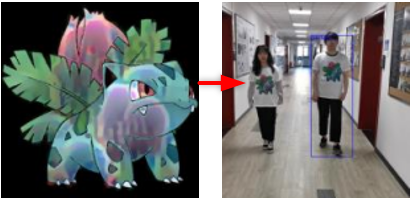}
	\end{minipage}
    \begin{minipage}{.49\linewidth}
    \centerline{\footnotesize Hu~\etal\cite{Hu2021NaturalisticPA}} 
    \centering
    \includegraphics[width =1\linewidth]{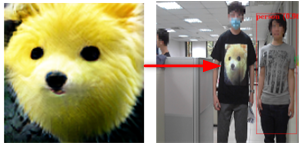}
	\end{minipage}
	\caption{Examples of Natural Physical Adversarial Attacks}
	\label{fig:natural_examples}
\end{figure}


Physical attacks tend to ignore the modification magnitude of adversarial patch/camouflage for better attack performance. However, a conspicuous perturbation would alert the potential victim and cause the attack to fail.
Therefore, research has focused on constructing inconspicuous perturbations in the real world to perform natural physical adversarial attacks, as shown in Fig.~\ref{fig:natural_examples}. 
The main techniques used can be categorized as optimization-oriented methods and generative model-oriented methods.
For clarity, we explicitly refer to optimization-oriented methods as optimizing for individual adversarial examples, and generative model-oriented methods as optimizing within the latent space of the data-driven generative models. 

\ding{182} \textbf{Optimized-oriented methods.}
The mainstream research on optimization-oriented methods for natural attacks has focused on how to define naturalness metrics for optimization that are practically meaningful, such as defining the metrics for optimizing invisibility, insignificance, and semantic naturalness.

Initially, researchers attempted to make adversarial patches look like a particular benign patch to expose the security problems of deep learning models.
Duan \etal~propose AdvCam~\cite{duan2020adversarial} that minimizes (1) the style distance $\mathcal{L}_s$ between the patch and the referenced image, (2) the content distance $\mathcal{L}_c$ between the patch and the background and (3) maximizes the smoothness loss defined by the total-variation loss $\mathcal{L}_{tv}$ in Eq.~\eqref{equ:tv-loss}.
The naturalness loss can be formalized as:
\begin{equation}
    \mathcal{L}_{nature} = \mathcal{L}_s + \mathcal{L}_c +\mathcal{L}_{tv}.
\end{equation}
Squared Euclidean distance on content features is applied as the content distance, and the 2-norm of the Gramian matrix of style features is applied as the style distance.
As a useful algebraic tool for style similarity~\cite{Gatys2016ImageST}, the Gramian matrix $\mathcal{G} = \{G_{i,j}\}$ is defined by $G_{ij} = f_i^Tf_j$, where $f_i$ and $f_j$ are feature vectors.
Then they use the gradient-based method to optimize the patch.
Much of the subsequent work\cite{huang2020universal,DBLP:conf/cvpr/WangLYLTL21} has taken a similar approach that adopts the natural image as the initialization of the adversarial patch, and then continuously optimizing the composed objective, which is generally a simple way to generate semantic natural patches.

Further works focus on defining specific natural style objectives. Tan \etal~\cite{tan2021legitimate} proposed \textit{LAP} to generate natural adversarial patches that evade the human eye. To achieve this goal, rationality loss is added to the patch training process that constrains the similarity of color features.
Cheng \etal~\cite{cheng2022physical} developed stealthy adversarial patch attacks against the monocular depth estimation model by minimizing the patch size and leveraging natural style. 
The author introduced neural style transfer to yield a natural style patch (e.g., rusty and dirty). 

Another perspective is to define a domain of natural adversarial examples in which to optimize. This may be more effective in concrete manufacturing.
Hu~\etal~\cite{hu2023physically} use discrete distribution on template colors and Voronoi diagram, implemented as the smoothed color distribution, to model the camouflage clothing, and perform the optimization on the distribution to generate adversarial camouflage.
As a general solution, Gumbel softmax technique is used to tackle the problem of non-differentiable optimization space caused by discrete distributions.

\ding{183} \textbf{Generative model-oriented methods.}
Deep generative methods aim to learn the distribution sampling function, or neural network, from the given dataset and can be used as an effective method for generating natural adversarial examples.
The advantage of generative models is that they reduce the difficulty of defining optimization objectives in conjunction with the learned data distribution.
In general, the generative method can be formulated as:
\begin{equation}
    \mathcal L_{natural} = \mathbb E_{x\sim P_{real}, y\sim P_{adv}}(\mathcal D(x, y)),
\end{equation}
where $x\sim P_{real}$ are real data sampled from the training dataset, $P_{adv}$ is the distribution generated by the attack model $G_{\theta}(P_{real})$, and $\mathcal D(\cdot, \cdot)$ is the pre-defined(in VAE models) or adversarially learned (in GAN models) distance metric. Specifically, $\mathcal D(x, y) = - \log(D_{\theta}(x)) - \log(1 - D_{\theta}(y))$ in the vanilla GAN, where $D_{\theta}(\cdot)$ is the adversarially trained discriminator network.

Early studies in the digital world have shown the effectiveness of GAN in adversarial attacks~\cite{2018AdvGAN}.
Liu~\etal~proposed \textit{PS-GAN}~\cite{Liu2019PerceptualSensitiveGF} that adapts GAN for generating less perceptually sensitive PAEs in the physical world.
The PS-GAN framework consists of a generator network that generates the adversarial patch from the seeded patch, and a discriminator that promises perceptual similarity between the image with PAE applied and the clean image.
Furthermore, an attention model is constructed with the clean image as input to determine the optimal location of the adversarial patch.
Bai \etal~\cite{bai2021inconspicuous} generated more inconspicuous adversarial patches that closely resembled background objects based on the saliency map and multiple generators and discriminators at different scales.
However, this approach require training a series of network pairs for each object sample, which can be resource intensive.

It is also feasible to generate PAEs in the latent space of existing well-trained generative networks, which indicates a broad distribution of PAEs.
Qiu \etal~\cite{qiu2020semanticadv} proposed \textit{SemanticAdv} to generate semantically realistic adversarial examples with an attribute-conditioned image editing network that can synthesize images with desired attributes.
Additionally, the method optimizes the adversarial example by interpolating semantically truthful feature spaces.
Hu \etal~\cite{Hu2021NaturalisticPA} proposed the method of generating natural PAEs inside the latent space of cycleGAN that pretrained in natural images.
Doan \cite{doan2022tnt} trained a generator to learn a mapping from a latent variable to a natural image of a flower.
Their framework also successfully generated transferable and generalizable adversarial patches.
To improve the naturalness of PAEs for real face recognition systems, Jia \etal~\cite{jia2022adv} developed a StyleGAN-based attack method involving high-level semantic perturbations.
The researchers altered facial characteristics to create more realistic and inconspicuous adversarial faces.

We can also model human visual perception.
To further address the inconsistency of evaluation of existing studies that mostly use subjective metrics, Li \etal~\cite{li2023towards} took the first step to benchmark and evaluate the naturalness of physical adversarial attacks. 
The authors constructed a Physical Attack Naturalness dataset that contains multiple attacking methods and human ratings, and a Dual Prior Alignment network is proposed that aligns human behavior with model decisions to automatically evaluate the naturalness of physical adversarial attacks. 
However, keeping the robustness and the generalization ability of the alignment network under the adversarial attack scenario requires further investigation.

\subsubsection{The Transferable Physical Adversarial Attacks}

The transferability of PAE measures whether the adversarial examples are highly aggressive across models.
Previous work on adversarial attacks in the digital world has shown that the same adversarial sample can exhibit generic attack capabilities for different deep learning models~\cite{DBLP:conf/iclr/LiuCLS17}.
This indicates the prevalence of vulnerable patterns across deep learning models.
To generate PAEs with transferability capability, the attack model needs to find these examples among these vulnerable patterns that can be implemented in the physical world.
These examples can further help to demonstrate the limits of the deep model's pattern recognition capabilities in the real world.
The potential transferability is an important property of adversarial examples, and the transferable properties of adversarial examples remain an open question even in the digital world.
Formally, referring to Eq.~\eqref{equ:pae}, for the generator $\delta(x)$ trained to maximize 
$\mathcal{D}(y^x ,  \mathcal{F}_1(x_{adv}^p)), \emph{s.t.} \lVert x_{adv}^p\rVert _{\aleph} < \varepsilon$, the scenario of transferable physical adversarial attacks requires that the adversarial example $\delta(x)$ be evaluated and tested on other models:
\begin{equation}
    \mathcal{D}(y^x ,  \mathcal{F}_2(x_{adv}^p)), \quad x_{adv}^p =x + \mathcal{R}(\mathcal{M}(\delta (x)), c),
\end{equation}
where $\mathcal{F}_1$ and $\mathcal{F}_2$ are different models.

In order to generate examples with high transferability when generating adversarial examples against a surrogate model, one heuristic is to target vulnerable features in the surrogate model that may be shared with other models, which can be located by the attention interpretation tools.
Wang \etal~\cite{DBLP:conf/cvpr/WangLYLTL21} suppress both model and human attention to generate transferable and natural adversarial camouflage based on the input seeded patch.
In order to distract model attention, they suppress the total attention score for each salient area in the model attention map, such as the result Grad-CAM~\cite{Selvaraju2016GradCAMVE}.
The results show satisfactory attack transferability in the physical world. 
Further, Zhang \etal~\cite{zhang2023boosting} focused on improving the transferability of physical attack by dispersing the model attention. However, unlike previous work on image classification, the author presented two improvements. First, the multiscale feature map was taken into account when designing the loss function; second, the attention map is smoothed by combining the feature map with respect to multiple transformed inputs.
Experimental results verify the effectiveness of their method in improving the physical adversarial attack.
Recently, Huang \etal~\cite{huang2023t} introduced a novel ShakeDrop module that randomly drops a subset of layers to generate massive variants of the detectors, which significantly increases the transferability of the adversarial patch.
This method is based on the intuition that, according to computational learning theory, the diversity of samples increases the transferability of the algorithm, while the transferability of adversarial examples comes from the diversity of the models attacked during training.

\subsubsection{The Generalized Physical Adversarial Attacks}

The attack performance of the PAE in different scenarios, or the generalization ability of physical adversarial attacks, is another key to studying the limitation of the deep learning models in the real world.
In general, the generalization ability over different target objects and different transformations are two important generalization problems to consider.
Formally, referring to Eq.~\eqref{equ:pae}, for the generator $\delta(x)$ trained to maximize 
$\mathcal{D}(y^x,  \mathcal{F}(x_{adv}^p)), \emph{s.t.} \lVert x_{adv}^p\rVert _{\aleph} < \varepsilon$ under the condition of:
\begin{equation}
    \quad x_{adv}^p =x + \mathcal{R}(\mathcal{M}(\delta (x)), c),\ x \sim X, \ c \sim C.
\end{equation}
The scenario of generalized physical adversarial attacks requires that the adversarial example $\delta(x)$ be evaluated in other data set and environment conditions, and tested in the condition of:
\begin{equation}
    \quad x_{adv}^p =x + \mathcal{R}(\mathcal{M}(\delta (x)), c),\ x \sim X', \ c \sim C',
\end{equation}
where $X$ and $X'$ are different data distributions, and $C$ and $C'$ are different environmental condition distributions.

The generalization ability of PAEs largely depends on the robustness of adversarial attacks against physical-to-digital transformation.
Wiyatno \etal~\cite{wiyatno2019physical} proposes the adversary for sequential tracking with the ability to attack in the physical world.
During the training they apply EoT on a mini-batch of object tracking scenes to enhance the robustness.
Experiments show that reducing the variance of target and background appearance in training does not strongly affect attack ability, while adding light transformation causes convergence difficulties.
In terms of limitations, attack training fails when the poster is too small.
In addition, Wu \etal~\cite{wu2020making} conducted a systematic study of the generalization ability of adversarial patch physical attacks on object detection and provided some interesting results: 1) The complex transformation operation such as thin-plate-spline (TPS) increases the training difficulty and then leads to a worse result in both the digital and physical world in some cases. 2) Physical attack performance is susceptible to imaging processes such as camera post-processing, pixelation, and compression.
What's more, since hand-designed transformations may not be able to fully cover the entire transformation space in the real world, Jan \etal~\cite{jan2019connecting} proposed a data-driven approach that uses a dataset to train a cycleGAN and uses it as a simulation of the digital-to-physical transformation.

To explore the class-agnostic generalized adversarial patterns, Liu and Wang \etal~\cite{DBLP:conf/eccv/LiuWLCZY20, wang2021universal} proposed a universal physical adversarial example with generalized attack capability for different classes.
The framework first trains a prior texture with high style similarity with hard examples, i.e., examples in the original dataset that are difficult to classify correctly, and with high attack capability that maximizes the uncertainty, measured by Shannon entropy, of the model output.
They use the Grad-CAM~\cite{Selvaraju2016GradCAMVE} method to measure and extract the region in the hard example that is significant to the model decision.
Then they fine-tuned the adversarial patch with a set of class prototypes, and showed that hard examples prior can improve adversarial patch generation.

For the vehicle recognition task, Suryanto \etal~\cite{suryanto2023active} proposed the Adversarial Camouflage for Transferable and Intensive Vehicle Evasion (ACTIVE) to generate universal and robust adversarial camouflage capable of concealing various vehicles against different detectors.
The tri-planar mapping (TPM) strategy, random output augmentation (ROA), and smooth loss are introduced into the training scheme for better robustness in real-world scenarios. In addition, the stealth loss and camouflage loss are proposed to promote generalization.

\section{Confront Physical Adversarial Examples}
Threatened by physical adversarial examples, it is necessary to investigate the solutions to protect intelligent applications from this kind of threat. Fortunately, there already exists a series of studies in enhancing the intelligent models and applications to defend against physical adversarial. We will give a detailed description of them in this section. Besides, it is also valuable to further analyze the future challenges and opportunities of the physical adversarial examples. More precisely, learning the challenges helps us generate stronger physical adversarial examples, which is beneficial to reveal the blind spots inside the models. Also, the physical adversarial examples will bring researchers fresh opportunities in multiple areas. For example, the physical adversarial examples might attach importance to re-consider the security issues, which gives bigger birth to the model robustness evaluation.
In this section, we first briefly introduce the hierarchy of the defense methods against the physical adversarial examples, including data-end defenses and model-end defenses. Moreover, we forecast the development of the physical adversarial examples research based on the aforementioned analysis and understanding that was proposed first in this paper. Finally, we aim to outlook the future research directions about the physical adversarial examples, pushing the negative influence away and encouraging the positive impacts.

\subsection{Defend against PAEs}
Since the physical adversarial examples pose significant threats to real-world applications, researchers have been making great efforts to defend against these dangerous examples to prevent models from murdering. Generally, a defense technique could be regarded and defined as a strategy that is able to alleviate the risks from both digital and physical attacks.  

Technically, digital adversarial defenses have limited differences from physical adversarial defenses. However, the physical world attacks urge the corresponding defenses to consider more about the real scenarios and applications. In this paper, a preferred perspective to categorize the adversarial defenses is to distinguish them via the object that the defending methods aim at. To be specific, the adversarial defenses could be categorized into data-end defenses and model-end defenses, the former aims to remove the adversarial noises or eliminate the effects of adversarial noises, and the latter aims to enhance the robustness of the models themselves.

\begin{figure}[b]
    \centering
    \includegraphics[width=.99\linewidth]{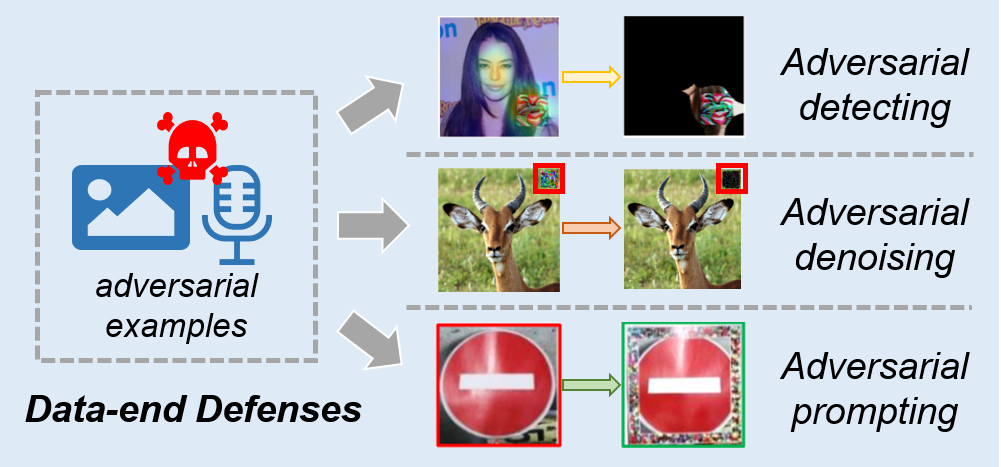}
    \caption{The sketch map of the data-end defenses.}
    \label{fig:data-end-defense}
\end{figure}

\begin{figure}
    \centering
    \includegraphics[width=.99\linewidth]{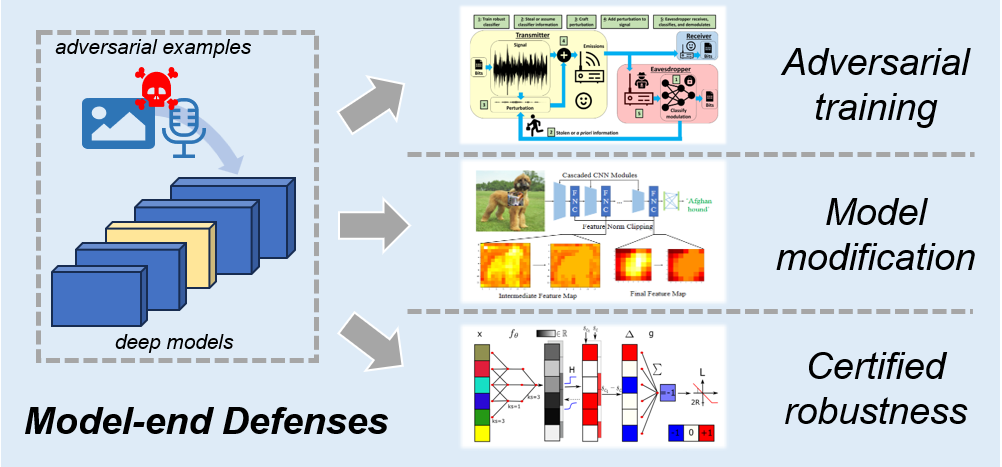}
    \caption{The sketch map of the model-end defenses.}
    \label{fig:model-end-defense}
\end{figure}

In detail, The data-end defense strategies aim to reduce the influence of adversarial perturbations, thus the sampled adversarial examples would not be allowed to mislead the deep models in deployed systems. In practice, we divide the data-end adversarial defenses into 3 categories that are composed of adversarial detecting, adversarial denoising, and adversarial prompting. The adversarial detecting methods mainly defend the PAEs by judging if the inputs are adversarial or not. These methods allow the AI systems to 
deny or locate the toxic parts. And the adversarial denoising methods could remove the injected perturbation or noises inside the PAEs so that the input instances could be benign again for model prediction. The adversarial prompting methods are also designed to inject additional information into PAEs, but could be able to offset the effectiveness of adversarial patterns.
The sketch map of the data-end defense can be found in Figure\ref{fig:data-end-defense}. 
The model-end defending methods are designed to improve the robustness of the models themselves. Therefore, the enhanced models could enjoy considerable performance when facing adversarial examples. There exists a taxonomy of model-end defense strategies, which includes adversarial training, structure improvement, and robustness certification. 
The adversarial training addresses the adversarial vulnerability problem by introducing the adversarial examples into the model training process, therefore helping the models acquire better generalization to the PAEs. 
The model modification methods defend the adversarial examples via modifying the model structures based on empirical observation or theoretical analysis. 
The certified robustness methods provide a theoretical robustness guarantee for deep models so that the users know how the model could be employed.
The sketch map of model-end defenses can be found in Figure\ref{fig:model-end-defense}. 

Due to the space limitation, we could not provide detailed introductions of current mainstream adversarial defenses. For more comprehensive descriptions, please refer to the Appendix, where we give a further statement of the existing popular and typical defense methods.

\subsection{The Challenges of PAEs}
Despite achieving great developments in the physical adversarial examples, we could still observe that there exists room for further improvement. As we mentioned before, the physical adversarial examples mainly focus on 3 kinds of attributes, \ie, basic attributes, core attributes, and epitaxial attributes, including 6 different attribute terms, \ie, naturalness, aggressiveness, re-sampled noises, re-sampler quality, transferability, and generalization. Up to now, a series of studies makes efforts on the basic attributes and core attributes, such as \cite{qiu2020semanticadv,duan2020adversarial,huang2020universal,DBLP:conf/cvpr/WangLYLTL21,tan2021legitimate,Liu2019PerceptualSensitiveGF,Selvaraju2016GradCAMVE,wiyatno2019physical,jan2019connecting,wu2020making}, while the transferable and generalizable attacking ability of physical adversarial examples remains in a relatively early stage. That means it is still a challenge to investigate the generation methods of physical adversarial examples that with higher transferability and generalization.
\subsubsection{Generating Transferable PAEs}
Since the physical adversarial examples play an impact in real scenarios, it is possible to face a situation that contains multiple target models. The case in point is that a gas station might have different surveillance cameras that carry different models. Thus, taking the transferable attacking ability of physical adversarial examples into account is reasonable. However, current transferable physical adversarial attack studies show significant insufficiency. Only a few works have been made on this topic, for instance, Wang \etal~\cite{DBLP:conf/cvpr/WangLYLTL21} try to generate transferable adversarial camouflage on image recognition tasks. Zhang \etal~propose a transferable attacking framework in the physical world to mislead detectors. 

The core problem in generating transferable physical adversarial examples is to capture the invariant space that is depicted together by different models. Deep models that are trained on various datasets might learn distinctive manifold space due to the distinction of the model architectures. However, they still show connective behavior performance, which indicates the shared or similar perception of feature space. Thus, generating physical adversarial examples targeted to these invariant spaces could be very beneficial to transferable attack ability. Unfortunately, constrained by the limited knowledge of the interpretability of deep models, it is not so easy to achieve this goal, making it a great challenge.

\subsubsection{Generating Generalizable PAEs}

Beyond the transferable physical adversarial examples, another challenge also comes from the complex and open environment but concentrates on the data end, \ie, the insufficient attacking ability to diverse classes and multiple instances. To be specific, generating different adversarial examples that correspond to different classes is impossible in physical attacking scenarios because of the uncountable and unpredictable objects in the real world. Thus, it is necessary to explore the generalizable generation method of physical adversarial examples to perform attacks efficiently and practically. Up to now, most of the investigation of generalizable adversarial attacking studies concentrates on the digital world rather than the physical world, which could not fulfill the requirements in practice. In open and complex scenarios, a huge amount of instances is sampled from different classes, which makes it difficult to generate physical adversarial examples case by case. And what's worse, there also exists noises like rain, snow, and fog, which results in distribution shifts and leads to a miserable attacking condition in turn. 

The critical keys to tackling these issues consist of two possible approaches that correlate to the intrinsic characteristics of adversarial examples themselves. One is to enhance the adversarial features contained in adversarial noises, thus the adversarial examples could capture and mislead the model's attention. Another is to enhance the perturbing ability to object patterns, which makes the models fail to recognize. Overall, the generalizable physical adversarial example generation is still at the first step, which is challenging and worth studying.

\subsection{The Opportunities of PAEs}
Though showing great threats from several aspects, the PAEs are not harmful only. Since the model robustness problems will not disappear because of overlooks, the occurrence of physical adversarial examples knolls the alarm bell of deep models' security and safety and brings fresh air to the related research areas. Here, we list the future opportunities of physical adversarial examples from 2 aspects.
\subsubsection{Evaluate the Application Robustness via PAEs}
Generally, the physical adversarial examples are designed to perform attacks in real-world scenarios, which means, it is possible to explore the blind spots and bugs of deep models by regarding the physical adversarial examples as test cases before their deployment. 
Currently, the physical adversarial examples mainly aim at strong attacking ability. However, as an evaluation approach, we do not need to pursue extremely high attacking strength. Serving as test instances, it is also valuable to investigate some of the core attributes mentioned in the Section. B. For example, taking the generalized physical adversarial examples could better help test out the general robustness of the deployed models in applications. Also, taking the natural physical adversarial examples as test cases will be beneficial to announce the worst robustness of the deployed deep models. Considering the core attributes, \ie, re-ampler noises and re-sampler quality allows the physical adversarial examples to be closer to the real scenarios, therefore we could introduce 3D environments and the generated test instances together to simulate the real conditions for evaluation, which is able to perform cost-effective assessment instead of performing large-scale real condition assessments, in some test-expensive tasks, like auto-driving. We appeal to the researcher to investigate more diverse physical adversarial examples generation methods as testing approaches, pushing it a bigger step to the positive side for social good.

\subsubsection{Protect the Application Privacy via PAEs}
Although the physical adversarial examples are designed to mislead deep models, it is possible to take them in a ``dog-eat-dog'' approach, \ie, employing the adversarial examples to defend against privacy leakage issues, including personal information, painting style, and so on. In general, due to the prediction-fooling ability of the physical adversarial examples, we can take them as defenders to protect the private information inside sampled examples from stealing. The case in point is playing the adversarial audio pieces in the physical world to evade the potential dialogue leakage caused by automatic eavesdropping devices, \ie, the deep models could not recognize the recorded audio pieces well so that the private information could remain safe. 
Also, mankind could also wear the adversarial T-shirt or paint the adversarial textures and camouflage on their clothing or accessories as well, thus making them not available to Malicious surveillance cameras, \ie, protect privacy. Moreover, it is worth investigating broader physical adversarial examples based on various materials, which could strongly support the kind of defense ability against negative machine intelligence in many more scenarios.
To sum up, protecting the application privacy via physical adversarial examples is not only practical but also meaningful since it provides an opportunity to re-understand the role of the adversarial examples, and even artificial intelligence techniques, in daily life. To be specific, when used positively, it leads to positive when used evilly, it leads to evil, therefore encouraging us to think more positively about utilizing adversarial examples.

Beyond the aforementioned application, we believe that the PAEs could be more widely employed in broader areas, such as auto-driving, security inspection, and social media. For instance, the vision PAEs could serve as a corner case in testing to discover potential flaws in autonomous driving perception models. The audio PAEs might help the car owner from being eavesdropped by illegal vehicle intelligence applications. In summary, different kinds of PAEs can play different positive roles in many scenarios that can be imagined and unexpected. We have great confidence in its role for social good. 

\section{Conclusions}

In this paper, we focus on the physical adversarial example studies and construct a novel hierarchy and a conceptual system for better understanding the PAEs, and promoting the development of practical adversarial attacks and defenses in turn. In detail, considering our concluded particularities and the attributes of PAEs, we classify the PAEs (\ie, the physical adversarial attacks) into manufacturing-oriented attacks, re-sampling-oriented attacks, and other attacks. For defending the PAEs, we combine the current adversarial defense methods and divide them into data-end defenses and model-end defenses. Upon the constructed hierarchy and system, we give a summary of the challenges and opportunities of PAEs. To be specific, we believe the transferable physical attack and the generalized physical attack could be great challenges in future studies. Also, we suggest paying more attention to delivering more positive perspectives in exploiting PAEs, such as introducing them to robustness evaluation and privacy protection. We believe that this paper could help researchers further realize the threats and potentials of the physical world adversarial examples, therefore better go closer to the future practical and trustworthy deep learning.

\section*{Acknowledgments}
We thank Haojie Hao and Zhengquan Sun for their efforts in constructing the sustainable open-source project of this paper. Click \href{https://github.com/jiakaiwangCN/awesome-physical-adversarial-examples}{\textcolor{pink}{here}} to visit our GitHub project.

\bibliographystyle{IEEEtran}
\bibliography{ref}

\clearpage

\appendices
\section{Adversarial Defense Strategies}
As we mentioned in the main paper, since the physical adversarial examples pose significant threats to real-world applications, researchers have been making great efforts to defend against these dangerous examples to prevent models from murdering. Generally, a defense technique could be regarded and defined as a strategy that is able to alleviate the risks from both digital and physical attacks.  Technically, digital adversarial defenses have limited differences from physical adversarial defenses. However, the physical world attacks urge the corresponding defenses to consider more about the real scenarios and applications. In this paper, a preferred perspective to categorize the adversarial defenses is to distinguish them via the object that the defending methods aim at. To be specific, the adversarial defenses could be categorized into data-end defenses and model-end defenses, the former aims to remove the adversarial noises or eliminate the effects of adversarial noises, and the latter aims to enhance the robustness of the models themselves.

\subsection{Data-end Defense Strategies}
The data-end defense strategies aim to reduce the influence of adversarial perturbations, thus the sampled adversarial examples would be not allowed to mislead the deep models in deployed systems. In practice, we divide the data-end adversarial defenses into 3 categories that are composed of adversarial denoising, adversarial detecting, and adversarial prompting.

\ding{182} \textbf{\emph{The adversarial detecting}}. Some researchers are committed to determining whether the input instances are adversarial. After discovering the adversarial examples, the models are allowed to reject these inputs and evade the attack in turn. 

Chou \etal~proposed the SentiNet to detect the universal adversarial patches by leveraging the susceptibility of models to attacks \cite{DBLP:conf/sp/ChouTP20}. The SentiNet is able to judge if an input instance is adversarial or not without any model modifications or architecture changes, which makes it practical in real scenarios. 
Ji \etal~proposed a plugin defense component based on the YOLO architecture, so-called Ad-YOLO, to detect the possibly existing adversarial patches \cite{ji2021adversarial}. They simply introduce an additional class label ``patch'' and achieve significant improvements when facing adversarial patches compared with naive YOLO. 
Li \etal~designed a novel adversarial detecting method, named TaintRadar, to detect the localized adversarial examples based on the fact that the critical regions would arise larger label variance \cite{li2021detecting}. It is reasonable that if the region lies an adversarial patch, then the model-predicted labels will show large differences after moving that region out. They conduct extensive experiments in both the digital and physical worlds, demonstrating its effectiveness.
Liu \etal~detect the adversarial patches in a segmentation way \cite{liu2022segment}. They train a patch segmentor to obtain a basic mask and then conduct shape completion to acquire the patch mask. In this way, the adversarial patches could be detected, and they could remove the regions corresponding to the mask and re-detect the benign objects.
\cite{kim2022defending} proposes an adversarial patch-feature energy-driven method to remove the deep characteristics of adversarial patches, protecting the detection models from being misled by adversarial patches. 
PatchZero aims to first detect the adversarial patches and then zero out them so as to evade the influence from adversarial textures \cite{DBLP:conf/wacv/XuXZCN23}.

\ding{183} \textbf{\emph{The adversarial denoising}}. This kind of defense method prevents models from being fooled by adversarial attacks at the instance level, \ie, straightly removing the injected perturbation or noises inside the adversarial examples. This kind of defense could also combine with the aforementioned adversarial detecting strategy, leading to better defending ability.

Nasser \etal~proposed the Local Gradient Smoothing (LGS) method to defend against the basic physical attacks, \eg, Localized and Visible Adversarial Noise (LaVAN) and Adversarial patch, in an earlier time \cite{DBLP:conf/wacv/NaseerKP19}. The LGS will first estimate the region with the highest probability of adversarial noises, and then it enables the gradient activity of the explored corresponding regions to be reduced. Thus, the adversarial examples could be recognized by models correctly.
McCoyd \etal~proposed an interesting approach to cast off the influence from adversarial patches by partially occluding the image around each candidate patch location, so that the adversarial patches might be destroyed by these occlusions \cite{mccoyd2020minority}. Since this defense path depends on occluding patches, it could be regarded as a special implementation of denoising, \ie, casting off the influence of adversarial noises. 
\cite{DBLP:conf/mm/ChiangCW21} proposes the adversarial pixel masking method (APM) to defend against the physical attacks, \eg, adversarial patches. The core process of the APM is to train an adversarial pixel mask module to remove the adversarial patches according to the obtained adversarial pixel mask. 

The mentioned PatchZero in the adversarial detecting part could also be regarded as a denoising strategy \cite{DBLP:conf/wacv/XuXZCN23}, which in fact is a combination of adversarial detecting and denoising. It clearly indicates that adversarial detecting and denoising could be employed to tackle adversarial attacks simultaneously.  

\ding{184} \textbf{\emph{The adversarial prompting}}. This kind of defense achieves the defending target by contributing additional information to offset the negative impacts of adversarial perturbations, \ie, prompting what labels the models should truly predict via positive injections.

In \cite{DBLP:conf/nips/SalmanIEVMK21}, Salman \etal~first propose the unadversarial example, which conducts defense via generating the textures with prompting ability, in a 3D environment, \ie, a simulated physical scenario. They successfully generate the so-called ``robust object'' based on the input-perturbation-sensitivity of deep models, drawing a new map for physical adversarial defenses. 
Moon \etal~propose a prompting-like preemptive robustification approach to defend against the potential intercept-and-perturb behavior in real scenarios \cite{moon2022preemptive}. The proposed method allows the discovery of robust perturbations under a bi-level optimization scheme that could be added to images.
Similar to \cite{DBLP:conf/nips/SalmanIEVMK21}, Wang \etal~proposed the defensive patch to help models recognize the images easier by pre-injecting different positive patches into instances \cite{DBLP:conf/cvpr/WangYHLTQLT22}. Specifically, they take a further step to enhance the prompting intensity, \ie, injecting strong global perceptual correlations and local identifiable patterns, compared with the \cite{DBLP:conf/nips/SalmanIEVMK21}, leading to an awesome defending ability to both adversarial patches and common corruption.
Based on a viewpoint of the underlying manifold, \cite{kim2023amicable} tries to generate a kind of visual prompting perturbations, named ``amicable aid'', which are even able to provide a universal improvement for classification.
Chen \etal~also pay attention to the effectiveness of adversarial visual prompting \cite{chen2023visual}. They investigate the non-effectiveness of universal visual prompting and propose class-wise adversarial visual prompting instead.
Si \etal~investigate the angelic patch, which is also a kind of visual adversarial prompting, to enhance the detecting ability of detectors against distribution shifts and synthetic perturbations \cite{si2023angelic}.



\subsection{Model-end Defense Strategies}
The model-end defending methods are designed to improve the robustness of the models themselves. Therefore, the enhanced models could enjoy considerable performance when facing adversarial examples. In theory, the model-end adversarial defense methods could be regarded as a more realistic approach compared with data-end ones, because the former fundamentally solves the problem of model adversarial vulnerability. Also, there exists a taxonomy of model-end defense strategies, which includes adversarial training, structure improvement, and robustness certification.

\ding{182} \textbf{\emph{Adversarial training}}. In practice, adversarial training methods are the most pragmatic approaches to defending adversarial attacks \cite{madry2017towards,zhang2021interpreting,liu2021training,liu2023towards}. It allows the deep models to learn more robust decision boundaries so that the adversarial inputs will not be misclassified.

Rao \etal{} investigate the adversarial robustness of models to adversarial patches by firstly designing a location-optimized adversarial patches generation approach \cite{DBLP:conf/eccv/RaoSS20}. By adopting typical adversarial training with the generated adversarial patches, they could effectively improve the model robustness and even clean accuracy. 
Wu \etal{} adopt a similar adversarial training approach with Rao \etal{} for defending against the adversarial patches but with the help of the proposed attacking method Rectangular Occlusion Attacks (ROA) \cite{DBLP:conf/iclr/WuTV20}. Compared with \cite{DBLP:conf/eccv/RaoSS20},  \cite{DBLP:conf/iclr/WuTV20} considers more about the physical constraints and gives a more comprehensive analysis.
Matzen \etal{} proposed a meta adversarial training strategy, which introduces the meta learning into the adversarial training scheme for overcoming the patch changes during training, to defend against the universal adversarial patches in computer vision tasks, including image classification and traffic-light detection \cite{metzen2021meta}.
Considering the huge practical threats in the real world, McClintick \etal{} turn their gaze to the wireless
communication scenarios and propose a feasible approach to improve the robustness of radio frequency machine learning signal classification models with realistic settings, \ie, designing adversarial training scheme \cite{9916315}. Different from the existing studies in this area, they conduct both simulated and physical demonstrations to comprehensively support the effectiveness of the proposed defense research.
Taking monocular depth estimation (MDE) as a typical and security-critical task, \cite{DBLP:conf/iclr/0010LTL023} proposed the view synthesis-based adversarial training strategy with full exploitation of the MDE's domain knowledge, \eg, physical-world constraints. By introducing the $l_{0}$-norm-bounded perturbation into the training scheme,  they achieve significant robustness improvement in defending against physical attacks.

\ding{183} \textbf{\emph{Model modification}}. This kind of defense mainly improves the model's robustness by modifying the model structures based on empirical observation or theoretical analysis \cite{liu2023exploring,tang2021robustart,xie2019denoising}. In this way, the models could achieve more stable performance under the challenge of adversarial examples.

BlurNet is a typical study that aims at defending against adversarial patches like robust physical perturbation ($RP^2$) from the perspective of frequency space \cite{DBLP:conf/dsn/RajuL20}. It removes the high-frequency noise from the adversarial examples by simply introducing a low-pass filter layer after the first layer of the model, achieving considerable defending ability.
Xiang \etal~proposed the PatchGuard to defend against the adversarial patches by introducing the small receptive fields as fundamental blocks for models \cite{xiang2021patchguard}, achieving high provable robust accuracy value and clean accuracy value. 
Motivated by the observation that the large norms caused by universal adversarial patches, \cite{DBLP:conf/iccv/YuCXLWBM21} proposes a feature
norm clipping (FNC) module, which could be flexibly inserted into any different models that are based on convolutional neural networks. 
Motivated by the evidence accumulation mechanism, Chen \etal~proposed the Dropout-based Drift-Diffusion Model (DDDM) to promote the robust performance of deep models \cite{DBLP:conf/ijcai/ChenLZY22}. The brain mechanism-based design makes the DDDM enjoy strong flexibility to multiple tasks, \ie, task-agnostic.
\cite{DBLP:journals/remotesensing/WangWZL23} first proposed the robust feature extraction network (RFENet) to address the adversarial vulnerability problem for aerial image
semantic segmentation task. Specifically, the RFENet consists of a limited receptive field mechanism (LRFM), a spatial semantic enhancement module (SSEM), a boundary feature perception module (BFPM), and a global correlation encoder module (GCEM), leading to considerable robustness against the adversarial patch attacks.
Kang \etal~proposed a diffusion-based adversarial defense strategy, which is allowed to locate and restore the adversarial patches \cite{DBLP:journals/corr/abs-2306-09124}. Besides the common image classification task, they also conduct demonstrations on the face recognition task, making the results reliable.
Based on the observation that adversarial patches always attract attention, Yu \etal~pay their attention to improving the robustness of vision transformers (ViTs) against adversarial patches \cite{yu2023defending}, achieving certain results supported by the validations on ImageNet.
Deb \etal~proposed a unified detection framework to defend against multiple attacks, including digital and physical adversarial attacks, based on a multitask learning paradigm \cite{DBLP:conf/fgr/DebLJ23a}. The well-trained end-to-end model could detect attack behaviors, \eg, predict if the input is an adversarial input or not, and even further classify what attack it is, \ie, adversarial, digitally manipulated, or contains spoof artifacts.

\ding{184} \textbf{\emph{Certified robustness}}. The certified defenses are proposed to provide a theoretical robustness guarantee for deep models. Given the guaranteed robustness bound, we acquire the robustness ability, which represents the ability to resist the $\mathcal{L}_p$-constrained attacks, of a certain model.

Chiang \etal~proposed the first certified defense strategy against adversarial patches with considerable training speed \cite{DBLP:conf/iclr/ChiangNAZSG20}. By introducing an interval bound, they construct a modified certification problem for patch attacks. Also, the \emph{random patch certificate training} and \emph{guided patch certificate training} are designed to speed up the training. 
Zhang \etal~proposed a so-called clipped BagNet (CBN) scheme to defend against adversarial stickers (patches) and they gave a certified security guarantee for the proposed CBN \cite{DBLP:conf/sp/ZhangYMW20}.
For achieving more practical robustness for safety-critical domains, Hendrik \etal~propose the BAGCERT that allows the trained models to enjoy better-certified robustness performance, \ie, higher clean accuracy, higher certified accuracy, and faster efficient inference \cite{DBLP:conf/iclr/MetzenY21}. The proposed BAGCERT is composed of a novel certifying condition, an updated model architecture, and an end-to-end training procedure. 
\cite{DBLP:conf/ccs/0001M21} proposes a provably securing object detectors by designing an objectness explaining strategy. The key idea lies on mismatch judgment, which is provided by an objectness explainer, of the objectness predictor and base detector.
Aiming at the same issue, Chen \etal~introduced the Vision Transformer architecture into the Derandomized Smoothing framework and achieved superior certifiable patch defense in both clean and certified accuracy \cite{Chen_2022_CVPR}. 
\cite{DBLP:conf/uss/0001MM22} proposes a double-masking defense framework, PatchCleanser, to perform considerable certifiable robustness against adversarial patches for image classification. The PatchCleanser efficiently alleviates the high dependence on model architectures and is allowed to make predictions facing adversarial inputs.
Based on the fact that the most aforementioned certified robustness enhancement for classification, Yatsura \etal~proposed the first certified defense against for semantic segmentation models via the designed de-masked smoothing \cite{DBLP:conf/iclr/YatsuraSH0M23}.

\section{The Paper List of the Physical Adversarial Attacks and Defenses}
The paper list of the \textit{Manufacturing Process Oriented PAE Methods} is shown in Tabel~\ref{tab:manufacture}, the \textit{Re-sampling Process Oriented PAE Methods} is shown in Table~\ref{tab:resample}, the \textit{``Other'' PAE Methods} is shown in Tabel~\ref{tab:other_paes}, and the \textit{Defense Methods against PAE} is shown in Tabel~\ref{tab:defensetable}.


\begin{table*}
\centering
\footnotesize
\caption{The overview of ``Manufacturing Process Oriented'' physical adversarial examples.}
\resizebox{\linewidth}{!}{
\begin{tabular}{llllllll}
\toprule\hline\noalign{\smallskip}
Categories &       & Method                                     & Publication           					& Year & Source & Task & Carrier \\\midrule\hline\noalign{\smallskip}
Touchable  & 2D    & Sharif \etal~\cite{sharif2016accessorize}               & ACM CCS						    & 2016 & Not available & Face Recognition & Eyeglasses\\\cline{3-8}\noalign{\smallskip}
		   &       & Lu \etal~\cite{lu2017adversarial}                   & ArXiv                             & 2017 & Not Available & Object Detection & Road Sign \\\cline{3-8}\noalign{\smallskip}
           &       & Kurakin \etal~\cite{ifgsm2018adversarial}                & ICLR                              & 2018 & Not Available & Image Classification & Poster \\\cline{3-8}\noalign{\smallskip}

           &       & Eykholt \etal~\cite{eykholt2018robust}                   & CVPR						        & 2018 & Not Available & Image Classification & Road Sign \\\cline{3-8}\noalign{\smallskip}                                 
           &       & Eykholt \etal~\cite{eykholt2018physical}                 & USENIX Workshop                   & 2018 & Not Available & Object Detection & Road Sign \\\cline{3-8}\noalign{\smallskip}
           &       & Lee \etal~\cite{lee2019physical}                     & ICML Workshop			            & 2019 & Available & Object Detection & Patch \\\cline{3-8}\noalign{\smallskip}
            &       & Thys \etal~\cite{Thys2019fooling}                     & CVPR Workshop			            & 2019 & \href{https://gitlab.com/EAVISE/adversarial-yolo}{Available}  & Object Detection & Patch \\\cline{3-8}\noalign{\smallskip}
           &       & Sharif \etal~\cite{sharif2019general}                   & ACM TOPS                          & 2019 & \href{https://github.com/mahmoods01/agns}{Available} & Face Recognition & Eyeglasses \\\cline{3-8}\noalign{\smallskip}
		   &       & Wang \etal~\cite{wang2021daedalus}                    & TCYB                              & 2021 & \href{https://github.com/NeuralSec/Daedalus-attack}{Available} & Object Detection & Patch \\\cline{3-8}\noalign{\smallskip}		   
     		           
           &       & Wei \etal~\cite{wei2022adversarial}                  & TPAMI                             & 2022 & \href{https://github.com/jinyugy21/Adv-Stickers_RHDE}{Available} & Face Recognition & Patch \\\cline{3-8}\noalign{\smallskip}  
           &       & Wei \etal~\cite{wei2022simultaneously}               & TPAMI                             & 2022 & \href{https://github.com/shighghyujie/newpatch-rl}{Available} & Face Recognition & Patch \\\cline{3-8}\noalign{\smallskip}
           &       & Wei \etal~\cite{wei2023hotcold}                      & AAAI                              & 2023 & \href{https://github.com/weihui1308/HOTCOLDBlock}{Available} & Object Detection & Paste  \\\cline{3-8}\noalign{\smallskip}
           &       & Wei \etal~\cite{wei2023physically}                   & CVPR                              & 2023 & \href{https://github.com/shighghyujie/infrared_patch_attack}{Available} & Object Detection & Paste \\\cline{3-8}\noalign{\smallskip}
          &       & Wei \etal~\cite{wei2023unified}                   & ICCV                              & 2023 & \href{https://github.com/Aries-iai/Cross-modal_Patch_Attack}{Available} & Object Detection & Paste \\\cline{3-8}\noalign{\smallskip}
            &       & Zhang \etal~\cite{zhang23CAPatch}                   & USENIX Security Symposium                              & 2023 & \href{https://github.com/USSLab/CAPatch}{Available} & Image Captioning & Patch \\\cline{2-8}\noalign{\smallskip}
           &  3D     & Athalye \etal~\cite{athalye2018synthesizing}             & ICML                              & 2018 & \href{https://github.com/prabhant/synthesizing-robust-adversarial-examples}{Available} & Image Classification & 3D Object \\\cline{3-8}\noalign{\smallskip}
           &       & Xiao \etal~\cite{xiao2019meshadv}                     & CVPR   							& 2019 & Not Available & \makecell[l]{Object Detection \&  Image Classification}  & 3D Object \\\cline{3-8}\noalign{\smallskip}
           &       & Xu \etal~\cite{xu2020adversarial}           		& ECCV                     & 2020 & Not Available & Object Detection & Clothes \\\cline{3-8}\noalign{\smallskip}    
           &       & Tu \etal~\cite{tu2020physically}           		& CVPR                    & 2020 & Not Available & Object Detection & 3D Object \\\cline{3-8}\noalign{\smallskip}             
           &       &Huang~\etal~\cite{huang2020universal}           	& CVPR               & 2020 & \href{https://github.com/mesunhlf/UPC-tf}{Available} & Detection & 3D Object \\\cline{3-8}\noalign{\smallskip} 
           &       & Wang \etal~\cite{DBLP:conf/cvpr/WangLYLTL21}                        & CVPR   							& 2021 & \href{https://github.com/nlsde-safety-team/DualAttentionAttack}{Available} & Object Detection & 3D Texture \\\cline{3-8}\noalign{\smallskip}
           &       & Wang \etal~\cite{wang2022fca}           				& AAAI                  		    & 2022 & \href{https://github.com/winterwindwang/Full-coverage-camouflage-adversarial-attack}{Available} & Object Detection & 3D Texture \\\cline{3-8}\noalign{\smallskip}

           &       & Miao \etal~\cite{miao2022Isometric3DAdv}           			& NIPS     & 2022 & Not available & Object Recognition & 3D Object \\\cline{3-8}\noalign{\smallskip}
           &       &Suryanto~\etal \cite{suryanto2022dta}           			& CVPR                  		    & 2022 & Available &\\\cline{3-8}\noalign{\smallskip}  
           &       &Zhang~\etal \cite{zhang2023boosting}       			&CVPR                                & 2023 & Available &\\\cline{3-8}\noalign{\smallskip}  
           &       & Hu \etal~\cite{hu2023physically}           			& CVPR                  			& 2023 & Not Available & Object Detection & Clothes \\\cline{3-8}\noalign{\smallskip} 
           &       & Yang \etal~\cite{yang2023towards}       				& CVPR                              & 2023 & Available & Face Recognition & 3D Object \\\cline{3-8}\noalign{\smallskip}  
           &       & Liu \etal~\cite{liu2023x}       				        & USENIX Security Symposium         & 2023 & \href{https://github.com/DIG-Beihang/X-adv}{Available} & Object Detection & 3D Object \\\cline{3-8}\noalign{\smallskip} 

            &       & Chen \etal~\cite{chen2024local}       				        & Computers \& Security       & 2024 & \href{https://github.com/Chenfeng1271/L3A}{Available} & Point Cloud Classification & 3D Object \\\cline{3-8}\noalign{\smallskip} 
           &       & Lou \etal~\cite{lou2024hide}       				        & CVPR         & 2024 & \href{https://github.com/TRLou/HiT-ADV}{Available} &  Classification   & 3D Object \\\midrule\hline\noalign{\smallskip}

            Untouchable  & Lighting  & Nichols \etal~\cite{nichols2018projecting}      & AAAI Symposium                          & 2018 & Not available  & Image Classification & Projector \\\cline{3-8}\noalign{\smallskip}
             &           & Man \etal~\cite{man2019poster}              & IEEE S\&P  & 2019 & Not available  & \makecell[l]{Object Detection \& \\ Image Classification} & Projector \\\cline{3-8}\noalign{\smallskip}
             &           & Guo \etal~\cite{guo2020watch}               & NeurIPS                                 & 2020 & \href{https://github.com/tsingqguo/ABBA}{Available} & Image Classification & Blur      \\\cline{3-8}\noalign{\smallskip}
             &           & Kim\etal~\cite{kim2021light}               & ArXiv                                   & 2021 & Not available & Image Classification & Modulator  \\\cline{3-8}\noalign{\smallskip}
             &           & Sayles \etal~\cite{sayles2021invisible}    	 & CVPR                                    & 2021 & \href{https://github.com/earlence-security/invis-perturbations}{Available}  & Image Classification & LED    \\\cline{3-8}\noalign{\smallskip}                        
             &           & Zhu \etal~\cite{zhu2021fooling}             & AAAI                                    & 2021 & Not available & Object Detection & Bulb \\\cline{3-8}\noalign{\smallskip}
             &           & Duan \etal~\cite{Duan2021AdversarialLB}      & CVPR                                    & 2021 & \href{https://github.com/RjDuan/Advlight}{Available}  & Image Classification & Laser     \\\cline{3-8}\noalign{\smallskip}
             &           & Pony \etal~\cite{pony2021over}               & CVPR                                    & 2021 & Not available & Video Recognition & LED  \\\cline{3-8}\noalign{\smallskip}
             &           & Gnanasambandam \etal~\cite{gnanasambandam2021optical}  & ICCV Workshop                           & 2021 & Not available & Image Classification & Projector  \\\cline{3-8}\noalign{\smallskip}            
             &           & Lovisotto \etal~\cite{lovisotto2021slap}          & USENIX Security Symposium  			   & 2021 & \href{https://github.com/ssloxford/short-lived-adversarial-perturbations}{Available} & Image Classification & Projector \\\cline{3-8}\noalign{\smallskip}            
             &           & Speth \etal~\cite{speth2022digital}           & WACV             					   & 2022 & Not available & Pulse Detection & LED \\\cline{3-8}\noalign{\smallskip}
             &           & Huang \etal~\cite{huang2022spaa}              & VR               					   & 2022 & \href{https://github.com/BingyaoHuang/SPAA}{Available}  & Image Classification & Projector  \\\cline{3-8}\noalign{\smallskip}
             &           & Hu \etal~\cite{hu2022advlen}               & ArXiv              					   & 2022 & Not available & Image Classification & Lens\\\cline{3-8}\noalign{\smallskip}
             &           & Hu \etal~\cite{hu2022advfilm}              & ArXiv              					   & 2022 & Not available & Image Classification & Sheet\\\cline{3-8}\noalign{\smallskip}
             &           & Zhong \etal~\cite{zhong2022shadows}           & CVPR                                    & 2022 & \href{https://github.com/hncszyq/ShadowAttack}{Available} & Image Classification & Shadow \\\cline{3-8}\noalign{\smallskip}
             &           & Li \etal~\cite{li2023physical}             & CVPR                                    & 2023 & Not available & Face Recognition & Fringe \\\cline{3-8}\noalign{\smallskip}
             &           & Wang \etal~\cite{wang2023rfla}               & ICCV                                    & 2023  & \href{https://github.com/winterwindwang/RFLA}{Available} & Image Classification & Reflection\\\cline{3-8}\noalign{\smallskip} 
             & & Li \etal~\cite{schmalfuss2023distracting}             & ICCV                                    & 2023 & \href{https://github.com/cv-stuttgart/DistractingDownpour}{Available} & Motion Estimation  & Weather Effects\\\cline{3-8}\noalign{\smallskip}   
            & & Hsiao \etal~\cite{hsiao2024natural}             & CVPR                                    & 2024 & \href{https://github.com/BlueDyee/natural-light-attack}{Available} & Road Sign Classification  & Natural Light\\\cline{3-8}\noalign{\smallskip}   
             
             & & Hu \etal~\cite{hu2024adversarialneon}           & Computer Vision and Image Understanding     & 2024 & \href{https://github.com/ChengYinHu/https://github.com/ChengYinHu/AdvNB}{Available} & Image Classification  &  Neon Beam Parameter Formalization\\\cline{2-8}\noalign{\smallskip}
             &  Audio/Speech    & Szurley \etal~\cite{szurley2019perceptual}  & ArXiv                                     & 2019 & Not available & Speech Recognition & Noise \\\cline{3-8}\noalign{\smallskip}
             &               & Qin \etal~\cite{qin2019imperceptible}   & ICML                                      & 2019 & \href{https://github.com/yaq007/cleverhans/tree/master/examples/adversarial_asr}{Available} & Speech Recognition & Noise \\\cline{3-8}\noalign{\smallskip}             
             &               & Yakura \etal~\cite{yakura2019robust}       & IJCAI                                     & 2019  & \href{https://github.com/hiromu/robust_audio_ae}{Available} & Speech Recognition & Noise\\\cline{3-8}\noalign{\smallskip}
             &               & Li \etal~\cite{li2019adversarial}      & NeurIPS                                   & 2019  & Not available & Speech Recognition & Music \\\cline{3-8}\noalign{\smallskip}
             &               & Chen \etal~\cite{chen2020devil}          & USENIX Security Symposium                 & 2020  & \href{https://github.com/RiskySignal/Devil-Whisper-Attack}{Available} &Speech Recognition & Noise \\\cline{3-8}\noalign{\smallskip}
             &               & Li \etal~\cite{li2020practical}        & HotMobile                                 & 2020  & Not available & Speaker Recognition & Noise\\\cline{3-8}\noalign{\smallskip}
             &               & Xie \etal~\cite{xie2020real}            & ICASSP              					     & 2020  & Not available & Speaker Recognition & Noise\\\cline{3-8}\noalign{\smallskip}
             &               & Sugawara \etal~\cite{sugawara2020light}      & USENIX Security Symposium                 & 2020  & Not available & Voice Conversion & Laser \\\cline{3-8}\noalign{\smallskip}             
             &               & Shang \etal~\cite{zhang2021attack}        & ICASSP                                    & 2021  & \href{https://github.com/zhang-wy15/Attack_practical_asv}{Available} & Speaker Verification & Noise\\\cline{3-8}\noalign{\smallskip}             
             &               & O'Reilly \etal~\cite{o2022voiceblock}        & NeruIPS                					 & 2022  & \href{https://github.com/voiceboxneurips/voicebox}{Available} & Speaker Recognition & Noise\\\cline{3-8}\noalign{\smallskip}          
             &   & Wang \etal~\cite{wang2022phonemic}       & Arxiv & 2022 & \href{https://anonymous.4open.science/r/Phonemic-Adversarial-Attack-1135/README.md}{Available} & Speech Recognition & Noise \\\cline{3-8}\noalign{\smallskip}                
             &               & Chiquier \etal~\cite{chiquier2022realtime}   & ICLR                 					 & 2022  & \href{https://github.com/cvlab-columbia/voicecamo}{Available} & Speech Recognition & Noise\\\bottomrule\hline\noalign{\smallskip}
\end{tabular}}
\label{tab:manufacture}
\end{table*}


\begin{table*}
\centering
\caption{The overview of ``Re-sampling Process Oriented'' physical adversarial examples.}
\resizebox{\linewidth}{!}{
\begin{tabular}{lllllll}
\toprule\hline\noalign{\smallskip}
Categories & Method                                        & Publication           				& Year & Source & Concrete re-sample approach & Task \\\midrule\hline\noalign{\smallskip}
Environment-oriented Attacks   
		 & Qin \etal~\cite{qin2019imperceptible}           & ICML						        & 2019 & \href{https://github.com/yaq007/cleverhans/tree/master/examples/adversarial_asr}{Available} & Reverberation by Room Impulse Response (RIR) & Speech Recognition\\\cline{2-7}\noalign{\smallskip}  
		 & Adversarial Music \cite{li2019adversarial}      & NeurIPS						    & 2019 & Not Available & Reverberation by Room Impulse Response (RIR) & Speech Recognition\\\cline{2-7}\noalign{\smallskip}  
		 & Li \etal~\cite{li2020practical}                 & HotMobile						    & 2020 & Not Available & Reverberation by Room Impulse Response (RIR) &Speech Recognition\\\cline{2-7}\noalign{\smallskip}  
		 & Xie \etal~\cite{xie2020real}                    & ICASSP						        & 2020 & Not Available & Reverberation by Room Impulse Response (RIR) & Speech Recognition\\\cline{2-7}\noalign{\smallskip} 
		 & Zhang \etal~\cite{zhang2021attack}              & ICASSP						        & 2021 & \href{https://github.com/zhang-wy15/Attack_practical_asv}{Available}     & Reverberation by Room Impulse Response (RIR)& Speech Recognition\\\cline{2-7}\noalign{\smallskip}  
		 & Du \etal~\cite{Du2021PhysicalAA}                & WACV						        & 2021 & Not Available & Brightness, Contrast, color &Object Detection \\\cline{2-7}\noalign{\smallskip} 
		 & Ding \etal~\cite{ding2021towards}               & AAAI						        & 2021 & Not Available & Brightness, Contrast, color & Object Tracking \\\cline{2-7}\noalign{\smallskip} 
        & Sayles \etal~\cite{sayles2021invisible}    & CVPR		 & 2021 & \href{https://github.com/earlence-security/invis-perturbations}{Available} & Brightness, exposure, color &  Image Classification \\\cline{2-7}\noalign{\smallskip} 
		 & DTA \cite{suryanto2022dta}                      & CVPR						        & 2022 & Not Available & Weather condition (e.g., shadow) & Object Detection \\\cline{2-7}\noalign{\smallskip}  
       & Chen~\etal~\cite{chen2022advICS}   & TDSC   & 2022 & Not Available & Network communication restriction modeling & Intrusion Detection
Systems  \\\cline{2-7}\noalign{\smallskip}
& Chen~\etal~\cite{chen2024local}   & Computers \& Security   & 2024 & \href{https://github.com/Chenfeng1271/L3A}{Available} & Environments Simulation & Point Cloud Classification  \\\cline{2-7}\noalign{\smallskip}
    & LESSON~\cite{Tian2024}& TDSC   & 2024 & Not Available & Power grid simulation & Injection Attacks Detection 
    \\\midrule\hline\noalign{\smallskip}
Sampler-oriented Attacks 
        & EoT\cite{athalye2018synthesizing}                & ICML                          & 2018 & \href{https://github.com/prabhant/synthesizing-robust-adversarial-examples}{Available}      & Expectation over Transformation(EoT) & Image Classification \\\cline{2-7}\noalign{\smallskip} 
		& ShapeShifter\cite{chen2018shapeshifter}          & ECML-PKDD                     & 2018 & \href{https://github.com/shangtse/robust-physical-attack}{Available}      & Expectation over Transformation(EoT) & Object Detection \\\cline{2-7}\noalign{\smallskip} 
		& $RP_2$ \cite{eykholt2018robust}                  & CVPR                          & 2018 & \href{https://github.com/evtimovi/robust_physical_perturbations}{Available}  & Digital-to-Physical Modeling & Image Classification\\\cline{2-7}\noalign{\smallskip} 
		& Eykholt \etal~\cite{eykholt2018physical}         & USENIX Workshop               & 2018 & Not Available  & Digital-to-Physical Modeling & Object Detection\\\cline{2-7}\noalign{\smallskip} 
		& AdvPattern\cite{wang2019advpattern}              & ICCV                          & 2019 & \href{https://github.com/whuAdv/AdvPattern}{Available}  & Multi-view augmentation & Reidentification \\\cline{2-7}\noalign{\smallskip} 
		& CAMOU\cite{zhang2018camou}                       & ICLR                          & 2019 & Not Available  & Multi-view augmentation & Object Detection\\\cline{2-7}\noalign{\smallskip} 
		& Zeng \etal~\cite{zeng2019adversarial}            & CVPR                          & 2019 & Not Available  & Multi-view augmentation & Vision Question Answering\\\cline{2-7}\noalign{\smallskip} 
	    & Adversarial Music\cite{li2019adversarial}        & NeurIPS                       & 2019 & Not Available  & Auditory masking calculated by a psychoacoustic model& Wake-word Detection\\\cline{2-7}\noalign{\smallskip} 
		& Szurley \etal~\cite{szurley2019perceptual}       & ArXiv                         & 2019 & Not Available  & Auditory masking calculated by a psychoacoustic model& Speech Recognition\\\cline{2-7}\noalign{\smallskip} 
		& Qin \etal~\cite{qin2019imperceptible}            & ICML                          & 2019 & \href{https://github.com/yaq007/cleverhans/tree/master/examples/adversarial_asr}{Available}  & Auditory masking calculated by a psychoacoustic model& Speech Recognition\\\cline{2-7}\noalign{\smallskip} 
        & Camera Sticker\cite{li2019sticker}                & ICML                          & 2019 & Not Available  & Color augmentation& Image Classification\\\cline{2-7}\noalign{\smallskip} 
		& Tsai \etal~\cite{tsai2020robust}                 & AAAI                          & 2020 & \href{https://github.com/jinyier/ai_pointnet_attack}{Available}      & Multi-view augmentation & Point Cloud Classification\\\cline{2-7}\noalign{\smallskip} 
		& Adversarial T-shirt\cite{xu2020adversarial}      & ECCV                          & 2020 & \href{https://github.com/jandress94/adversarial_tshirt}{Available}      & Deformable transformation (TPS) &Object Detection\\\cline{2-7}\noalign{\smallskip} 
		& Translucent Patch \cite{zolfi2021translucent}    & CVPR                          & 2021 & Not Available  & Color augmentation& Object Detection\\\cline{2-7}\noalign{\smallskip}	
		& CAC \cite{duan2022learning}                      & AAAI                          & 2022 & Not Available  & Multi-view augmentation & Object Detection\\\cline{2-7}\noalign{\smallskip} 
		& Shapira \etal~\cite{shapira2022attacking}        & ArXiv                         & 2022 & Not Available  & Deformable transformation & Object Detection\\\cline{2-7}\noalign{\smallskip} 
		& Hu \etal~\cite{hu2022adversarial}            & CVPR                          & 2022 & \href{https://github.com/WhoTHU/Adversarial_Texture}{Available}  & Expectation over Transformation(EoT) & Object Detection\\\cline{2-7}\noalign{\smallskip} 		
		& Zhu \etal~\cite{zhu2022infrared}                 & CVPR                          & 2022 & Not Available  & Expectation over Transformation(EoT), TPS & Object Detection \\\cline{2-7}\noalign{\smallskip} 
		& ViewFool \cite{dong2022viewfool}       	       & NeurIPS                       & 2022 & \href{https://github.com/Heathcliff-saku/ViewFool_}{Available} & Multi-view augmentation & Image Classification  \\\cline{2-7}\noalign{\smallskip} 
  & Wang~\etal~\cite{wang2023does}       	       & ICCV                       & 2023 & Not Available & Autonomous driving system simulation & Object Detection   \\\cline{2-7}\noalign{\smallskip} 
   & Guesmi~\etal~\cite{guesmi2024dap}       	       & CVPR                       & 2024 & Not Available & Creases Transformation & Object Detection   \\\cline{2-7}\noalign{\smallskip} 
    & Hu~\etal~\cite{hu2024adversarialinfrared}       	       & Neural Networks                       & 2024 &  \href{https://github.com/ChengYinHu/AdvIB}{Available} & Multi-view Physical Parameters Optimization & Thermal Infrared Detection   \\\bottomrule 
\end{tabular}}
\label{tab:resample}
\end{table*}
\begin{table*}
\centering
\caption{The Overview of ``Other'' Physical Adversarial Examples (PAE).}
\resizebox{\linewidth}{!}{
\begin{tabular}{lllllll}
\toprule\hline\noalign{\smallskip}
Categories   & Reference                  & Publication          & Year            & Source &  Method & Task                                         
\\\midrule\hline\noalign{\smallskip}
            
Natural PAE
			& Liu~\etal~\cite{Liu2019PerceptualSensitiveGF}   & AAAI               & 2019 & \href{https://github.com/liuaishan/PerceptualSensitiveGAN}{Available} & Generative & Classification\\\cline{2-7}\noalign{\smallskip}
			& Qiu~\etal~\cite{qiu2020semanticadv}          	& ECCV               & 2020 & \href{https://github.com/AI-secure/SemanticAdv}{Available}& Generative & Classification \\\cline{2-7}\noalign{\smallskip} 
			& Duan~\etal~\cite{duan2020adversarial}         	& CVPR               & 2020 & \href{https://github.com/RjDuan/AdvCam-Hide-Adv-with-Natural-Styles}{Available} & Optimization-oriented & Classification \\\cline{2-7}\noalign{\smallskip} 
			& Huang~\etal~\cite{huang2020universal}           	& CVPR               & 2020 & \href{https://github.com/mesunhlf/UPC-tf}{Available} & Optimization-oriented & Detection \\\cline{2-7}\noalign{\smallskip} 			
			& Tan~\etal~\cite{tan2021legitimate}           	& ACM MM             & 2021 & Not Available & Optimization-oriented & Detection \\\cline{2-7}\noalign{\smallskip}
			& Bai~\etal~\cite{bai2021inconspicuous}           & IEEE IoT Journal   & 2021 & Not Available & Generative & Classification \\\cline{2-7}\noalign{\smallskip}			
            & Hu~\etal~\cite{Hu2021NaturalisticPA}           & ICCV               & 2021 & \href{https://github.com/aiiu-lab/naturalistic-adversarial-patch}{Available} &  Generative & Detection \\\cline{2-7}\noalign{\smallskip}
			& Cheng~\etal~\cite{cheng2022physical}           	& ECCV               & 2022 & Not Available & Optimization-oriented & Depth Estimation \\\cline{2-7}\noalign{\smallskip}
            & Doan~\etal~\cite{doan2022tnt}               	    & IEEE TIFS          & 2022 & Not Available  & Generative & Classification  \\\cline{2-7}\noalign{\smallskip} 
            & Jia \etal~\cite{jia2022adv}               		    & NeurIPS                           & 2022 & Not Available & Generative & Face Recognition\\ \cline{2-7}\noalign{\smallskip}
            & Hu \etal~\cite{hu2023physically}           			& CVPR                  			& 2023 & Not Available & Optimization-oriented & Detection\\ \cline{2-7}\noalign{\smallskip} 
            & Li~\etal~\cite{li2023towards}               	& CVPR               & 2023 & \href{https://github.com/zhangsn-19/PAN}{Available} & Generative & Naturalness Benchmark \\\midrule\hline\noalign{\smallskip}
Transferable PAE
      & Wang~\etal~\cite{DBLP:conf/cvpr/WangLYLTL21}                  		& CVPR 				  & 2021 & \href{https://github.com/nlsde-safety-team/DualAttentionAttack}{Available} & Optimization-oriented & Classification \& Detection \\\cline{2-7}\noalign{\smallskip} 
      & Doan~\etal~\cite{doan2022tnt}               	    & IEEE TIFS          & 2022 & Not Available  & Generative & Classification  \\\cline{2-7}\noalign{\smallskip} 
      & Zhang~\etal~\cite{zhang2023boosting}             		& Pattern Recognition & 2023 & \href{https://github.com/zhangyu13a/transPhyAtt}{Available}& Optimization-oriented & Detection \\\cline{2-7}\noalign{\smallskip} 
      & Jia~\etal~\cite{jia2024fooling}                           & TITS                & 2024 & Not Available  &  Optimization-oriented & Detection \\\cline{2-7}\noalign{\smallskip}
      & Huang~\etal~\cite{huang2023t}                           & CVPR                & 2023 & \href{https://github.com/vdigpku/t-sea}{Available} &  Optimization-oriented & Detection \\\midrule\hline\noalign{\smallskip} 
           
Generalized PAE 
	      & Wiyatno~\etal~\cite{wiyatno2019physical}             & ICCV            & 2019 & Not Available & Optimization-oriented &  Object Tracking \\\cline{2-7}\noalign{\smallskip}
	      & Jan~\etal~\cite{jan2019connecting}               & AAAI            & 2019 & \href{https://github.com/stevetkjan/Digital2Physicale }{Available} & Generative & 
Classification \\\cline{2-7}\noalign{\smallskip}
	      & Wu~\etal~\cite{wu2020making}                    & ECCV            & 2020 & Not Available & Optimization-oriented & Detection \\\cline{2-7}\noalign{\smallskip}
	      & Liu~\etal~\cite{DBLP:conf/eccv/LiuWLCZY20}       & ECCV            & 2020 & \href{https://github.com/liuaishan/ModelBiasedAttack}{Available}  & Generative & 
Classification \\\cline{2-7}\noalign{\smallskip}      
	      & Wang~\etal~\cite{wang2021universal}               & IEEE TIP            & 2021 & \href{https://github.com/nlsde-safety-team/PerceptualAttentionalBiasedAttack}{Available} & Generative & 
Classification \\\cline{2-7}\noalign{\smallskip} 
        & Doan~\etal~\cite{doan2022tnt}               	    & IEEE TIFS          & 2022 & Not Available  & Generative & Classification  \\\cline{2-7}\noalign{\smallskip} 
        & Suryanto~\etal~\cite{suryanto2023active}               & ICCV            & 2023 & \href{https://islab-ai.github.io/active-iccv2023/}{Available} & Optimization-oriented & Detection
        \\\bottomrule\hline\noalign{\smallskip}
\end{tabular}}
\label{tab:other_paes}
\end{table*}

\begin{table*}
\centering
\caption{The list of the surveyed publications about adversarial defense strategies against physical attacks.}
\resizebox{\linewidth}{!}{
\begin{tabular}{lllllll}
\toprule\hline\noalign{\smallskip}
Categories &   & Method    & Publication   & Year &  Source &  Task   \\\midrule\hline\noalign{\smallskip}
Data-end   & Adversarial detecting & \cite{DBLP:conf/sp/ChouTP20}   & IEEE S\&P Workshop & 2020 & Not Available &    Face Recognition  \\
\cline{3-7}\noalign{\smallskip}
    &   & \cite{ji2021adversarial} & Arxiv & 2021 & Not Available &  Human Detection  \\
 \cline{3-7}\noalign{\smallskip}
 &  &  \cite{li2021detecting}  & IEEE INFOCOM & 2021 & Not Available &   Scene Classification \& Face Recognition  \\
 \cline{3-7}\noalign{\smallskip}
 &   & \cite{liu2022segment}& CVPR  & 2022 &  \href{https://github.com/joellliu/SegmentAndComplete}{Available} &   Object Detection \\
 \cline{3-7}\noalign{\smallskip}
 &   & \cite{kim2022defending}   & ACM MM & 2022 & Not Available & Object Detection \\
 \cline{3-7}\noalign{\smallskip}
 &  &  \cite{DBLP:conf/wacv/XuXZCN23} & WACV  & 2023 & Not Available &  Image$/$Video Classification \& Object Detection \\
 \cline{2-7}\noalign{\smallskip}
 & Adversarial denoising & 
 \cite{DBLP:conf/wacv/NaseerKP19} & WACV  & 2019 & Not Available &  Image Classification    \\
 \cline{3-7}\noalign{\smallskip}
  &   & \cite{mccoyd2020minority}  & ACNS Workshop  & 2020 & Not Available &   Image Classification  \\
  \cline{3-7}\noalign{\smallskip}
&   & \cite{DBLP:conf/mm/ChiangCW21}  & ACM MM  & 2021 & Not Available & Object Detection \& Human Detection \\
 \cline{3-7}\noalign{\smallskip}
 &  & \cite{DBLP:conf/wacv/XuXZCN23} & WACV   & 2023 & Not Available &  Image$/$Video Classification \& Object Detection   \\
 \cline{2-7}\noalign{\smallskip}
 & Adversarial prompting &\cite{DBLP:conf/nips/SalmanIEVMK21}  & NIPS  & 2021 & \href{https://github.com/microsoft/unadversarial}{Available} &   Image Classification  \\
 \cline{3-7}\noalign{\smallskip}
&  & \cite{moon2022preemptive} & AAAI   & 2022 & \href{https://github.com/snu-mllab/preemptive_robustification}{Available}  &    Image Classification  \\
\cline{3-7}\noalign{\smallskip}
 &   & \cite{DBLP:conf/cvpr/WangYHLTQLT22}  & CVPR   & 2022 & \href{https://github.com/nlsde-safety-team/DefensivePatch}{Available} &  Image$/$Roadsign Classification  \\
 \cline{3-7}\noalign{\smallskip}
 &  & \cite{kim2023amicable} & ICASSP  & 2023 & Not Available &  Image Classification  \\
 \cline{3-7}\noalign{\smallskip}
 &  & \cite{chen2023visual} & ICASSP & 2023 & \href{https://github.com/Phoveran/vp-for-adversarial-robustness}{Available} &  Image Classification  \\
 \cline{3-7}\noalign{\smallskip}
  &  & \cite{si2023angelic} & CVPR  & 2023 & \href{https://github.com/averysi224/angelic_patches}{Available} &  Object Detection  \\
  \midrule\hline\noalign{\smallskip}
Model-end  & Adversarial training  & \cite{DBLP:conf/eccv/RaoSS20} & ECCV Workshop  & 2020 & \href{https://github.com/sukrutrao/adversarial-patch-training}{Available} &  Image$/$Roadsign Classification  \\
\cline{3-7}\noalign{\smallskip}
 &  & \cite{DBLP:conf/iclr/WuTV20}  & ICLR  & 2020 & \href{https://github.com/tongwu2020/phattacks}{Available} & Image Classification  \\
 \cline{3-7}\noalign{\smallskip}
& & \cite{metzen2021meta}   & ICML workshop & 2021 & \href{https://github.com/boschresearch/meta-adversarial-training}{Available} &   Image Classification \& Object Detection \\
\cline{3-7}\noalign{\smallskip}
  &   & \cite{DBLP:journals/ojcs/McClintickHFHW22}  & IEEE ComSoc & 2022 & Not Available &  Signal Classification \\
  \cline{3-7}\noalign{\smallskip}
 &   & \cite{DBLP:conf/iclr/0010LTL023}    & ICLR  & 2023 & \href{https://github.com/Bob-cheng/DepthModelHardening}{Available} &  Monocular Depth Estimation  \\
 \cline{3-7}\noalign{\smallskip}
 &   & \cite{DBLP:conf/dsn/RajuL20} & DSN Workshops  & 2020 & Not Available &  Traffic Sign Classification  \\
 \cline{2-7}\noalign{\smallskip}
 &  Model modification & \cite{xiang2021patchguard} & USENIX Security Symposium  & 2021 & \href{https://github.com/inspire-group/PatchGuard}{Available} &  Image Classification  \\
 \cline{3-7}\noalign{\smallskip} &  & \cite{DBLP:conf/iccv/YuCXLWBM21}  & ICCV  & 2021 & Not Available &   Image Classification \\
 \cline{3-7}\noalign{\smallskip}
  &  & \cite{DBLP:conf/ijcai/ChenLZY22}  & IJCAI  & 2022 & \href{ https://github.com/XiYuan68/DDDM}{Available} &  Image$/$Audio Classification  \\
  \cline{3-7}\noalign{\smallskip}
 & & \cite{DBLP:journals/remotesensing/WangWZL23}& Remote sensing & 2023 & \href{https://github.com/darkseid-arch/PatchRFENet}{Available(\textcolor{red}{\emph{empty now}})} & Aerial Image Semantic Segmentation   \\
 \cline{3-7}\noalign{\smallskip}
 &  & \cite{DBLP:journals/corr/abs-2306-09124} & Arxiv & 2023 & Not Available &  Image Classification \& Face Recognition  \\
 \cline{3-7}\noalign{\smallskip}
 &   & \cite{yu2023defending}  & ICASSP & 2023 & Not Available &   Image Classification \\
 \cline{3-7}\noalign{\smallskip}
  &   & \cite{DBLP:conf/fgr/DebLJ23a} & IEEE FG & 2023 & Not Available &    Face Recognition \\
  \cline{2-7}\noalign{\smallskip}
 & Certified robustness  & \cite{DBLP:conf/iclr/ChiangNAZSG20} & ICLR  & 2020 & \href{https://github.com/Ping-C/certifiedpatchdefense}{Available} & Image Classification  \\
 \cline{3-7}\noalign{\smallskip}
 &  & \cite{DBLP:conf/sp/ZhangYMW20} & IEEE S\&P Workshop & 2020 & Not Available &  Image Classification  \\
 \cline{3-7}\noalign{\smallskip}
 &  & \cite{DBLP:conf/iclr/MetzenY21}   & ICLR   & 2021 & Not Available &  Image Classification  \\
  \cline{3-7}\noalign{\smallskip}
&     & \cite{10.1145/3460120.3484757}  & ACM CCS & 2021 & \href{https://github.com/inspire-group/DetectorGuard}{Available} &  Object Detection  \\
\cline{3-7}\noalign{\smallskip}
 &   & \cite{Chen_2022_CVPR}  & CVPR  & 2022 & Not Available &  Image Classification \\
 \cline{3-7}\noalign{\smallskip}
  &    & \cite{DBLP:conf/uss/0001MM22}  & USENIX Security Symposium   & 2022 & \href{https://github.com/inspire-group/PatchCleanser}{Available} &  Image Classification  \\
  \cline{3-7}\noalign{\smallskip}
 &  & \cite{DBLP:conf/iclr/YatsuraSH0M23}  & ICLR   & 2023 & Not Available &  Semantic Segmentation \\
 \bottomrule\hline\noalign{\smallskip}
\end{tabular}}
\label{tab:defensetable}
\end{table*}

\end{document}